\documentclass[11pt,a4paper]{article}

\usepackage[a4paper,margin=1in]{geometry}
\usepackage[T1]{fontenc}
\usepackage[utf8]{inputenc}
\usepackage{lmodern}
\usepackage{microtype}
\usepackage{graphicx}
\usepackage{amsmath,amssymb}
\usepackage{booktabs}
\usepackage{enumitem}
\usepackage{xcolor}
\usepackage{float}
\usepackage{hyperref}
\hypersetup{
  colorlinks=true,
  linkcolor=blue!60!black,
  citecolor=blue!60!black,
  urlcolor=blue!60!black
}
\usepackage{array}
\newcolumntype{C}[1]{>{\centering\arraybackslash}p{#1}}
\newcolumntype{L}[1]{>{\raggedright\arraybackslash}p{#1}}

\title{The Shape of Data: Topology Meets Analytics\\
\large A Practical Introduction to Topological Analytics and the Stability Index (TSI) in Business
}

\author{Ioannis Diamantis\\
\small Department of Data Analytics and Digitalisation, Maastricht University\\
\small Maastricht, The Netherlands\\
\small \texttt{i.diamantis@maastrichtuniversity.nl}
}

\date{} 

\begin{document}
\maketitle

\begin{abstract}
Modern business and economic datasets often exhibit nonlinear, multi-scale structures that traditional linear tools under-represent. Topological Data Analysis (TDA) offers a geometric lens for uncovering robust patterns, such as connected components, loops and voids, across scales. This paper provides an intuitive, figure-driven introduction to persistent homology and a practical, reproducible TDA pipeline for applied analysts. Through comparative case studies in consumer behavior, equity markets (SAX/eSAX vs.\ TDA) and foreign exchange dynamics, we demonstrate how topological features can reveal segmentation patterns and structural relationships beyond classical statistical methods. We discuss methodological choices regarding distance metrics, complex construction and interpretation, and we introduce the \textit{Topological Stability Index} (TSI), a simple yet interpretable indicator of structural variability derived from persistence lifetimes. We conclude with practical guidelines for TDA implementation, visualization and communication in business and economic analytics.
\end{abstract}

\makeatletter
\renewcommand\@makefnmark{}  
\makeatother
\renewcommand\footnoterule{}
\footnotetext{\textbf{Keywords:} Topological Data Analysis, Persistent Homology, Structural Stability, Symbolic Time Series, SAX, eSAX, Business Analytics, Market Segmentation, Systemic Risk, Time Series Clustering, Geometric Data Analysis.\\
\textbf{MSC 2020:} 62R40, 55N31, 91B84, 62H30, 68T09.}

\section{Introduction: Why the Shape of Data Matters}

\subsection{Motivation}

Over the past two decades, data analytics has become central to decision-making in business, finance and economics. Companies now record enormous amounts of information, from customer interactions and online behavior to market prices, risk indicators and supply-chain activity. Yet most analyses still rely on linear tools such as correlations, regressions or principal-component analysis (PCA). These approaches describe how variables move together but say little about the overall structure or shape of the data.

In reality, many datasets are not flat clouds of points but have curves, clusters and cycles: consumers may form distinct yet overlapping communities; financial assets can move through repeating volatility patterns; and production networks may branch or reconnect in complex ways. Understanding these patterns requires methods that look beyond averages and straight-line relationships.

Topological Data Analysis (TDA) provides such a view. It treats data as a collection of points in a space where distances measure similarity. By gradually connecting nearby points and observing how the network of connections evolves, TDA reveals stable features; groups that remain connected, loops that persist and gaps that signal missing or redundant structure. These geometric summaries capture relationships that traditional tools overlook, while remaining interpretable in practical terms such as segmentation, seasonality or diversification.

\subsection{From Statistics to Geometry}

Traditional statistical and econometric models focus on \emph{parametric inference}, estimating relationships among variables under distributional assumptions. Even non-parametric or machine-learning approaches such as neural networks often obscure structure behind black-box mappings. In contrast, \emph{Topological Data Analysis} (TDA) adopts a geometric viewpoint: instead of modeling variables directly, it studies the \emph{shape} traced by data in a high-dimensional space. The central tool of TDA, \emph{persistent homology}, measures how topological features appear and disappear across scales, yielding a multi-resolution summary known as a \emph{barcode} or \emph{persistence diagram}. These objects quantify qualitative aspects, such as connectivity, cycles and cavities, in a way that is invariant to coordinate transformations and stable under noise.

This perspective has proven powerful in disciplines where structure is complex and noise pervasive. Early applications appeared in sensor networks, neuroscience, and biology \cite{Carlsson2009,EdelsbrunnerHarer2010,Ghrist2008}, revealing hidden organization in genetic, neural and spatial data. More recently, TDA has entered economics and finance, where market networks, correlations and behavioral data exhibit nonlinear interdependencies. Studies have shown that persistence diagrams can detect regime changes in equity markets, structural breaks in foreign-exchange dependencies, and emergent consumer clusters that escape classical clustering or PCA \cite{Wasserman2018}. However, such approaches remain under-represented in mainstream business analytics curricula and practice, partly due to the perceived mathematical barrier and the scarcity of accessible introductions.

\subsection{Why a Topological Lens for Business Analytics}

Business and financial systems are characterized by feedback loops, cyclical behavior, and structural dependencies, properties naturally expressed in topological language. Markets alternate between calm and turbulent phases, supply chains oscillate between shortage and surplus, and consumer preferences evolve through recurring trends. Linear correlation captures average co-movement but cannot describe the geometry of these dynamics. For example, two markets may exhibit near-zero correlation yet form a persistent topological loop in their joint state space, indicating a structured but phase-shifted relationship.

\smallbreak 

A topological framework provides several advantages:
\begin{itemize}
  \item \textbf{Model-agnostic structure discovery.} TDA uncovers patterns directly from pairwise distances, without assuming linearity or a specific functional form.
  \item \textbf{Multi-scale robustness.} Persistence summarizes patterns that survive across scales of similarity, filtering out noise and emphasizing stable structure.
  \item \textbf{Interpretability.} Features such as the birth and death of connected components have natural economic and managerial analogies: market cohesion, regime fragmentation or the merging of consumer segments.
  \item \textbf{Integration with existing analytics.} Persistence summaries can be combined with clustering, PCA or machine-learning models as geometric features or stability indicators.
\end{itemize}

From a decision-making perspective, this geometric approach complements rather than replaces conventional analytics. It enables managers to answer questions such as: ``How stable are our market clusters as we vary similarity thresholds?'' or ``Do consumer behaviors form cycles rather than static groups?'' Such insights enrich descriptive, diagnostic and even predictive analytics.

\subsection{Positioning and Contribution}

This paper serves a dual purpose. First, it provides an \emph{accessible, figure-driven survey} of TDA tailored to readers in data science, business analytics and finance. While the mathematical foundations are well established, expository accounts often assume a level of abstraction unfamiliar to practitioners. Here, we focus on intuition and visualization: point clouds, simplicial complexes, barcodes and persistence diagrams are introduced through minimal notation and concrete examples.

Second, we synthesize insights from several applied studies, including prior work on symbolic time series and consumer behavior, equity-market regimes using SAX/eSAX representations \footnote{
We apply Symbolic Aggregate approXimation (SAX) and its extension, eSAX, for discretizing time series into symbolic strings, enabling fast clustering and visual pattern comparison. SAX reduces each time series to a word representation based on mean values over fixed-length segments, while eSAX retains more structure by encoding max–min–mean sequences. For details, see \cite{Lin2003} for SAX and \cite{Lkhagva} for eSAX.} and exchange-rate dependencies analyzed through topological lenses, into a unified analytical framework. Together, these examples demonstrate how topology can reveal structural information complementary to traditional statistical measures. Beyond summarizing results, we generalize them theoretically by discussing the role of distance metrics, the choice of complex and the implications of persistence for portfolio diversification and systemic risk.

\subsection{Related Work and Research Gap}

The mathematical foundations of TDA are documented in seminal works such as Carlsson (2009) \cite{Carlsson2009}, Edelsbrunner and Harer (2010) \cite{EdelsbrunnerHarer2010}, and Ghrist (2008) \cite{Ghrist2008}, which established persistent homology as a tractable tool for exploring complex data. Over the past decade, applications have gradually extended to finance and economics \cite{HobbelhagenDiamantis2024, DeFavereauDiamantis2025}. Studies of financial networks have used TDA to analyze correlation structures, detect crisis dynamics and model contagion paths, most notably Gidea and Katz \cite{Gidea2018}, who employed persistence landscapes to identify early warning signals in market data. Beyond finance, TDA has been applied to a wide range of systems, from physical and networked structures \cite{Kramar2013, Aktas2019} to biological and medical datasets \cite{Camara2016, Crawford2019, Lee2021}, and even to large-scale information environments where topology serves as a paradigm for big-data analytics \cite{Berwald2018}.

Despite these advances, applications that translate TDA concepts into the language of business and economic analytics remain limited. In particular, few studies address how choices of distance metric, complex construction or persistence interpretation can be aligned with managerial or market-oriented questions. The present paper fills this gap by providing an accessible, figure-driven exposition of these methodological aspects and by illustrating their value through comparative case studies in consumer behavior, equity markets and currency co-movements.

\subsection{Structure of the Paper}

The remainder of this paper is organized as follows. Section~2 introduces the geometric intuition behind TDA, from data clouds and distance functions to simplicial complexes and filtrations. Section~3 explains persistent homology and its visualization through barcodes and persistence diagrams. Section~4 presents the full TDA pipeline and integration with standard analytical workflows. Section~5 compares case studies across consumer behavior, equities, and foreign-exchange markets, illustrating how persistent features capture evolving structures. Section~6 offers theoretical and methodological reflections on distance choice, complex selection, and a proposed topological stability index linking persistence to portfolio and systemic risk. Section~7 provides practical guidelines and reporting standards, and Section~8 concludes with outlooks for future research and applications in data-driven decision making.

\section{From Data Clouds to Shapes: Building Intuition}

Topological Data Analysis (TDA) begins with something deceptively simple: a collection of data points.
Each observation, whether a consumer’s feature vector, a time window of stock returns, or a country’s macroeconomic indicators, can be viewed as a point in a high-dimensional space.
The cloud of points formed by all such observations has a shape that reflects underlying relationships: dense regions reveal clusters of similar behavior, loops capture recurrent or cyclical patterns, and empty regions (or cavities) mark structural gaps in diversification or activity.
Rather than imposing algebraic or parametric models, TDA studies this shape directly through geometry and connectivity, i.e. how points group and link as the notion of similarity expands.

\paragraph{From data to point clouds.}
In practice, a dataset becomes a point cloud once each observation is represented as a vector of features or measurements.  
For instance, a consumer profile may correspond to standardized preference scores; a stock to a time window of normalized returns or volatility; and a country to macroeconomic indicators such as GDP, inflation and trade balance.  
The resulting set of vectors $\{x_1, x_2, \dots, x_n\} \subset \mathbb{R}^d$ forms the input for topological analysis.  
Different pre-processing choices, standardization, time-delay embedding or symbolic encoding, affect the geometry of this cloud and thus the topology that emerges.

\subsection{Distances and Similarity Choices}

Every topological analysis starts by defining how ``close'' two observations are.  
The choice of distance determines which features of the data are emphasized, and different business contexts call for different metrics.  

In practice, analysts can choose from several notions of distance, depending on the type of data and the relationships of interest.  
Below we review the most common choices; geometric, statistical, directional and temporal, highlighting when each is meaningful and where it can fail.

\paragraph{Euclidean distance.}
The most common choice, measuring straight-line difference between observations across features.  
It is appropriate when variables are commensurate (same scale and units), for example standardized survey responses or normalized sales indicators.  
However, in time-series contexts Euclidean distance can be misleading: two markets with similar shapes but shifted in phase (one lags the other) may appear far apart.

\paragraph{Correlation distance.}
Defined as $d_{ij} = \sqrt{2(1-\rho_{ij})}$, where $\rho_{ij}$ is the Pearson correlation between series~$i$ and~$j$.  
This metric captures co-movement rather than magnitude, making it useful for financial returns or other relative signals.  
It automatically removes scale effects but assumes linear relationships and stationarity, conditions often violated during structural breaks or crises.

\paragraph{Cosine and angular distances.}
When data are represented as feature vectors (for example, marketing profiles or word embeddings), cosine similarity measures the angle between vectors and is insensitive to absolute magnitude.  
It is therefore popular in text, recommendation, and customer-segmentation analytics, where direction matters more than length.

\paragraph{Dynamic and symbolic distances.}
For sequential data, dynamic time-warping (DTW) aligns similar patterns occurring at different speeds or phases.  
In symbolic representations such as SAX or eSAX \cite{Lin2003}, distances are computed between symbolic strings, reflecting coarse behavioral similarity.  
Symbolic metrics reduce noise and dimensionality, at the cost of abstraction: small local variations may be ignored.

It must be highlighted that while DTW is a powerful tool, it must be noted that it is technically a {\it pseudo-metric} because it often violates the triangle inequality. This technicality does not negate its utility, but applied analysts should be aware that it affects the theoretical stability proof of the persistence diagram. Our decision to use DTW is based on its superior empirical performance in discriminating temporal patterns within our case studies, which we prioritize for the business application.

\paragraph{Practical caveats.}
Before computing distances, analysts should consider:
\begin{itemize}[leftmargin=1.2em]
  \item \emph{Scaling:} features with different units or volatilities can dominate distances if not normalized;
  \item \emph{Seasonality and non-stationarity:} changing variance or mean levels distort similarity; detrending or rolling-window analysis mitigates this;
  \item \emph{Noise sensitivity:} small perturbations can affect pairwise distances; using rank-based or correlation metrics often stabilizes results.
\end{itemize}

Choosing an appropriate distance is thus the first modeling decision in TDA.  
The persistence diagrams that follow do not depend on how data are labeled, but they depend sensitively on how proximity is defined.

\subsection{From Points to Simplicial Complexes}

Once distances are defined, we connect nearby points to form higher-dimensional objects called \emph{simplices}.  
A pair of points forms an edge (a~1-simplex), three mutually connected points form a triangle (a~2-simplex), and so on.  
A collection of such simplices satisfying simple intersection rules is a \emph{simplicial complex}.  
It can be seen as a scaffold capturing how data points group and overlap.

\paragraph{Building blocks: simplices.}
A $0$-simplex is a point, a $1$-simplex is an edge connecting two points, a $2$-simplex is a filled triangle, and a $3$-simplex is a tetrahedron.  
By gluing such simplices along shared faces, one constructs a simplicial complex, that is, a combinatorial structure that encodes how data points interconnect across scales.

\begin{figure}[H]\centering
\includegraphics[width=0.95\textwidth]{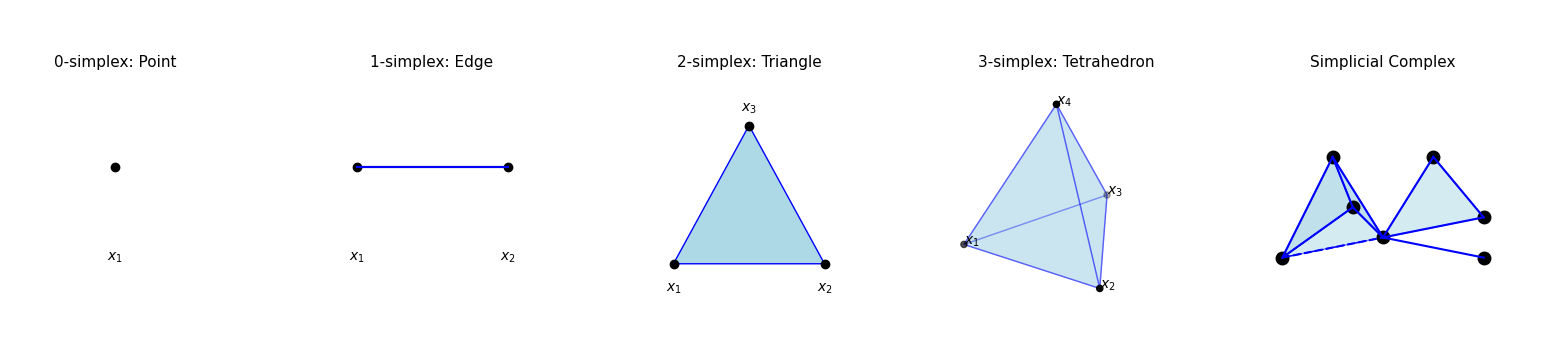}
\caption{Simplices of increasing dimension and their assembly into a simplicial complex. These geometric building blocks form the foundation of topological data analysis.}
\label{fig:simplices}
\end{figure}

Several constructions are common:

\paragraph{Vietoris–Rips complex.}
Connect every pair of points whose distance is below a threshold~$\varepsilon$; include all higher-dimensional simplices whose vertices are pairwise connected.  
This method depends only on pairwise distances and is computationally efficient.  
It tends to slightly over-estimate connectivity but is easy to compute for any metric space.

\paragraph{\v{C}ech complex.}
Imagine drawing a ball of radius~$\varepsilon/2$ around each point and connecting points whose balls intersect.  
This construction corresponds to the true underlying topology of the union of balls, but it requires knowledge of coordinates (not just distances) and is more expensive to compute.

The main conceptual difference between the Vietoris–Rips and \v{C}ech complexes lies in how they determine connectivity.
In the \v{C}ech complex, a simplex is formed when the corresponding balls (of radius~$\varepsilon/2$) centered at the points have a common intersection, that is, all points overlap in a single region.
In contrast, the Vietoris–Rips complex requires only that all pairwise distances between the points are within~$\varepsilon$.
As a result, the Rips complex is generally easier to compute and tends to be ``larger'' (containing more simplices) than the \v{C}ech complex at the same scale. It is worth mentioning that for any point cloud $X$ and any $\epsilon > 0$, the Rips complex $\text{Rips}_{\epsilon}(X)$ and the \v{C}ech complex $\check{C}_{\epsilon}(X)$ are related by the sequence of inclusions: $\check{C}_{\epsilon}(X) \subseteq \text{Rips}_{2\epsilon}(X)$

\begin{figure}[H]\centering
\includegraphics[width=0.6\textwidth]{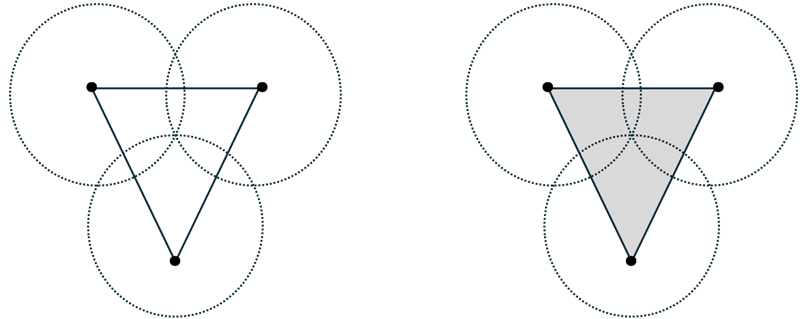}
\caption{Illustrative comparison of the \v{C}ech and Vietoris-Rips complexes.  
Left: in the \v{C}ech complex, a triangle appears only when the three corresponding balls have a common intersection.  
Right: in the Vietoris-Rips complex, the triangle is filled once all pairwise overlaps exist, even if there is no triple intersection.  
The Rips complex is therefore simpler to compute but slightly overestimates connectivity.}
\label{fig:cech-vs-rips}
\end{figure}

\paragraph{Witness complex.}
For large or high-dimensional datasets, computing all pairwise distances is prohibitive.  
Witness complexes use a subset of ``landmark'' points to approximate topology, drastically reducing computational cost while retaining essential structure.

\paragraph{Intuitive picture.}
At small~$\varepsilon$, every point is isolated.  
As~$\varepsilon$ grows, edges and triangles appear, and disconnected clusters begin to merge.  
At a critical scale, loops or voids may form; at larger scales, everything eventually becomes one connected component.  
Tracking these changes as the scale varies provides the essence of persistent homology.

\begin{figure}[H]\centering
\includegraphics[width=0.7\textwidth]{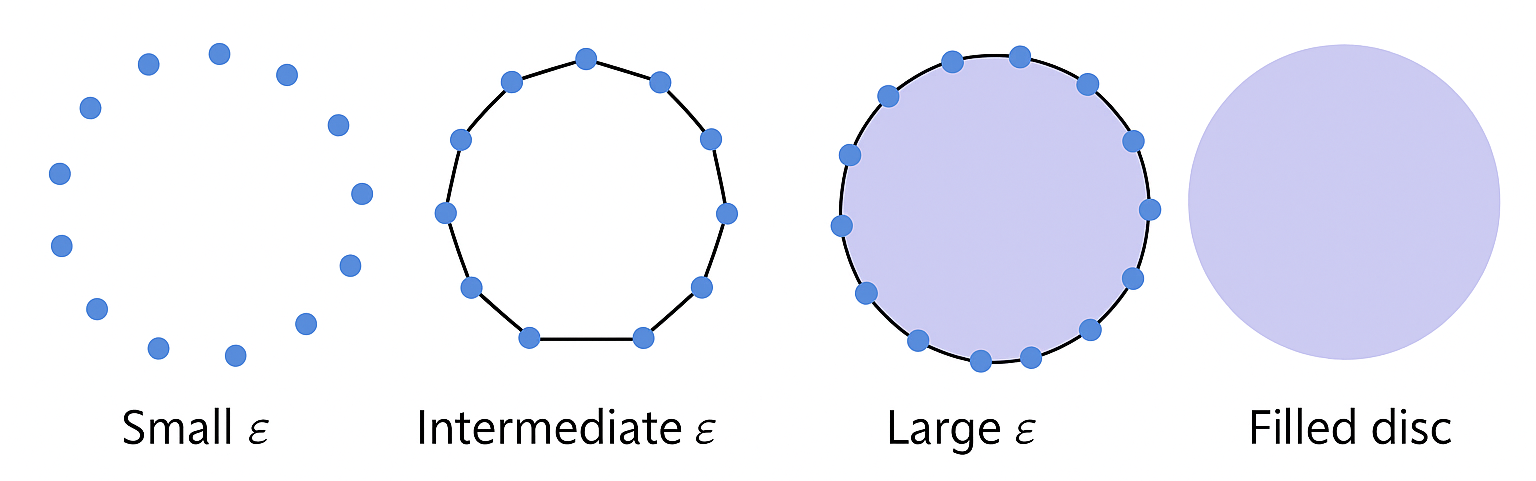}
\caption{Growth of a Rips complex on a circular point cloud as the distance threshold $\varepsilon$ increases. Small $\varepsilon$: isolated points; intermediate $\varepsilon$: connected ring; large $\varepsilon$: filled disc.}
\label{fig:rips-growth1}
\end{figure}

\subsection{Filtrations and Multi-Scale Thinking}

In practice, we do not fix a single threshold~$\varepsilon$ but examine a continuum of values.  
The sequence of nested complexes obtained as~$\varepsilon$ increases is called a \emph{filtration}.  
Filtrations are the backbone of TDA: they capture how topological structures emerge, merge and persist across scales.

From an analytical perspective, a filtration resembles a multi-resolution decomposition or a moving-average analysis:  
short-lived features correspond to noise, while long-lived ones reveal stable organization.  
For business analysts, this provides a natural notion of \emph{scale robustness}.  
If a consumer segment or market cluster persists as we vary the similarity threshold, it is likely a genuine feature of the system rather than an artifact of parameter choice.  
Likewise, if a loop (cycle) appears briefly and vanishes, it may correspond to transient or anomalous behavior.

\begin{figure}[H]\centering
\includegraphics[width=0.95\textwidth]{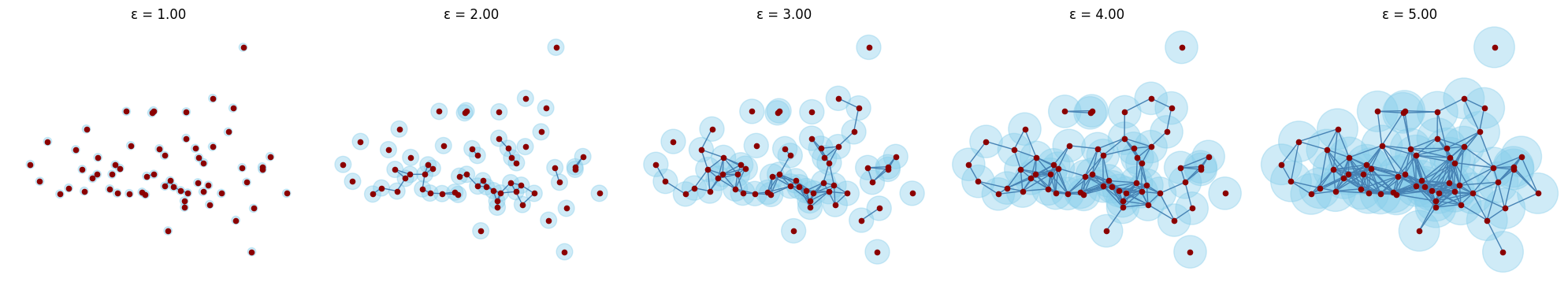}
\caption{Growth of a Vietoris--Rips complex as the distance threshold $\varepsilon$ increases. 
Small $\varepsilon$: isolated points; intermediate $\varepsilon$: edges and small clusters emerge; large $\varepsilon$: loops form and eventually fill in, connecting all points into a single component.}
\label{fig:rips-growth}
\end{figure}

By tracking which features are born and which survive longest along the filtration, we transform geometric growth into a compact summary of structure: a \emph{barcode} or \emph{persistence diagram}, introduced next.

\section{Persistent Homology Made Simple}

Persistent homology quantifies how connectivity evolves as we move through the filtration.  
It tracks the appearance (\emph{birth}) and disappearance (\emph{death}) of topological features at different scales, recording their lifetimes as signatures of structural persistence.

\subsection{Birth, Death and Persistence}

At the smallest scale, each data point is an isolated component, corresponding to a set of independent customers, stocks or countries.  
As $\varepsilon$ grows, points merge into clusters; each merger represents the death of a component in $H_0$.  
When loops or cycles form, for example, when three or more clusters connect around an empty region, a new feature in $H_1$ is born.  
If the loop later fills in, it dies.  
In higher dimensions, analogous ``voids'' (holes in 3D) correspond to $H_2$ and above.

The collection of all such births and deaths is captured in two equivalent visualizations:
\begin{itemize}
  \item \textbf{Barcodes:} horizontal segments whose lengths represent the lifetimes of features;
  \item \textbf{Persistence diagrams:} scatter plots where each feature is a point with coordinates (birth, death).
\end{itemize}

The longer a bar, or equivalently, the further a point lies from the diagonal, the more persistent the feature, and the more likely it reflects meaningful structure rather than noise.

\subsection{Barcodes and Persistence Diagrams}

A circular cloud of points produces many short $H_0$ bars that quickly merge into one component, plus one long $H_1$ bar representing the persistent loop.  
By contrast, a two-cluster dataset yields two long $H_0$ bars (the clusters) and no $H_1$ bar.  
Persistence thus distinguishes cyclic structure from mere grouping.

In applied contexts:
\begin{itemize}[leftmargin=1.2em]
  \item For consumer behavior, persistent $H_0$ bars may indicate stable preference segments;
  \item In markets, long $H_1$ bars can capture cyclical co-movement or alternating regimes;
  \item In portfolios, the emergence or disappearance of cycles may signal diversification or contagion shifts.
\end{itemize}

\begin{figure}[H]\centering
\includegraphics[width=0.65\textwidth]{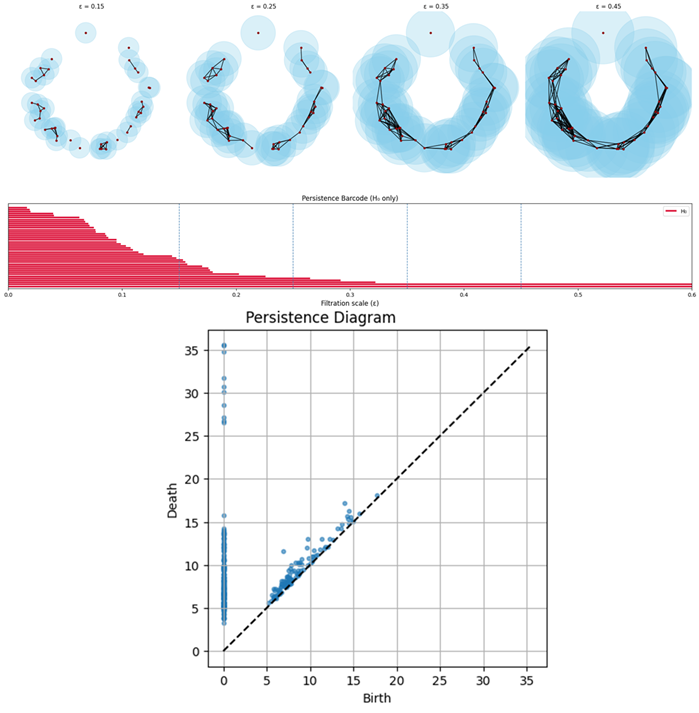}
\caption{Persistent homology visualized through complementary representations.  
Top: filtration of a circular dataset as the radius~$\varepsilon$ increases, points merge and a loop persists over several scales.  
Middle: persistence barcode showing feature lifetimes, where longer bars correspond to more stable structures.  
Bottom: persistence diagram plotting each feature by its birth and death coordinates; points farther from the diagonal represent robust, long-lived patterns.}
\label{fig:ph-combined}
\end{figure}

Persistence diagrams are particularly valuable because they are \emph{stable} under small perturbations~\cite{CohenSteiner2007}:  
if the underlying distances change slightly (due to measurement noise or rescaling), the diagram changes only slightly in the so-called bottleneck distance.  
This \emph{stability theorem} provides mathematical justification for using persistence as a reliable descriptor of structure in noisy empirical data.

\subsection{Interpretation for Non-Mathematicians}

Persistent homology \footnote{See \S~\ref{homtheory} if interested in the algebraic underpinnings of persistent homology.}
 provides a vocabulary for describing the \emph{shape} of data.  
Rather than focusing on numerical coordinates or statistical distributions, it summarizes how observations connect across scales.  
For analysts, long-lived components ($H_0$) reveal enduring groupings, while long-lived loops ($H_1$) capture cyclical or feedback behavior.  
The absence of persistent features implies either strong homogeneity (everything connects quickly) or extreme fragmentation (no connections form).

In marketing, this analysis can reveal when consumer clusters merge as similarity thresholds loosen; an indicator of overlapping interests.  
In finance, it can identify cycles in the joint movement of assets, corresponding to alternating risk regimes.  
In operations, loops may reflect recurring process states, while voids correspond to unvisited or infeasible configurations.

The strength of TDA lies in its universality: the same mathematical machinery applies to any domain where a notion of distance exists.  
By translating datasets into geometric objects and measuring their evolving shape, analysts gain an additional dimension of insight, not in variables, but in relationships.

\section{The TDA Pipeline for Analysts}

Having introduced the basic concepts of filtration and persistence, we now turn to the full analytical workflow that connects raw data to interpretable topological summaries.

The strength of Topological Data Analysis lies not only in its mathematical foundation but also in its reproducible workflow.  
Unlike many statistical models that depend on specific distributional assumptions, the TDA pipeline can be applied to virtually any dataset once a notion of distance or similarity is defined.  
This section outlines the full end-to-end process and interprets each step through the lens of business and economic analytics. Note that various efficient algorithms and practical considerations for computing persistent homology are surveyed in a ‘roadmap’ in \cite{Otter2017}.

\subsection{Overview}

\begin{figure}[H]\centering
\includegraphics[width=0.7\textwidth]{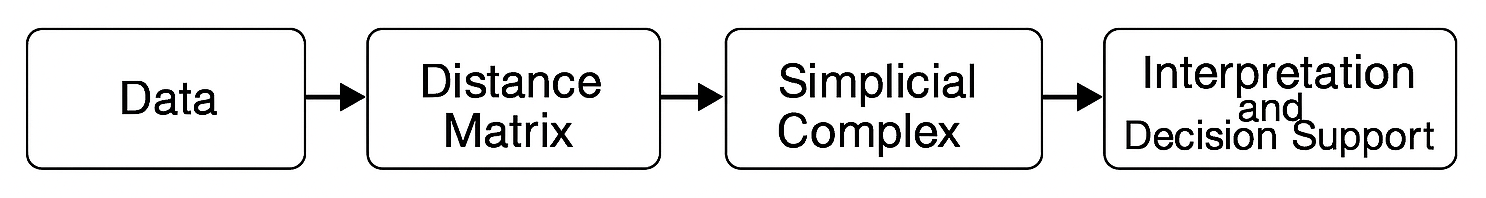}
\caption{Conceptual workflow of the TDA pipeline: raw data $\rightarrow$ distance matrix $\rightarrow$ simplicial complex and persistence computation $\rightarrow$ interpretation and decision support.}
\label{fig:pipeline}
\end{figure}

Figure~\ref{fig:pipeline} provides a schematic overview.  
Starting from raw data, one first constructs a distance matrix representing pairwise similarities.  
From these distances, one builds a simplicial complex that captures the data’s geometry.  
Persistent homology is then computed to summarize multi-scale topological features in the form of barcodes or persistence diagrams.  
Finally, these summaries are interpreted quantitatively or visually and integrated into downstream analyses such as clustering, forecasting or risk assessment.

\subsection{Step 1: Data Preprocessing}

The quality of any topological analysis depends on meaningful pairwise comparisons.  
Raw business data are often heterogeneous: variables may have different units, trends and seasonal patterns.  
Preprocessing therefore aims to create a homogeneous representation suitable for distance computation.  

Typical operations include:
\begin{itemize}[leftmargin=1.2em]
  \item \textbf{Normalization and scaling:} ensuring comparable ranges across variables, e.g., z-score standardization or log returns for financial time series;
  \item \textbf{Detrending and de-seasonalization:} removing long-term drift or periodic effects that can dominate distances;
  \item \textbf{Windowing:} transforming a long time series into overlapping segments (sliding windows) to study temporal evolution;
  \item \textbf{Noise reduction:} smoothing or symbolic representation (SAX, eSAX) to suppress high-frequency fluctuations.
\end{itemize}

These steps ensure that geometric distances capture genuine similarity rather than artefacts of scale, trend or noise.

\subsection{Step 2: Distance Matrix Computation}

Once pre-processing is complete, the next step is to compute the pairwise distance matrix $D=(d_{ij})$.  
Each entry quantifies dissimilarity between observations $i$ and $j$ using the metric most appropriate for the domain:
\begin{itemize}[leftmargin=1.2em]
  \item Euclidean distance for standardized cross-sectional data;
  \item Correlation or cosine distance for co-movement in financial or behavioral time series;
  \item Dynamic Time Warping (DTW) or symbol-based distances for temporal alignment;
  \item Domain-specific metrics (e.g., Wasserstein distance for distributions or histograms).
\end{itemize}

In TDA, the distance matrix is the only required input; it defines a metric space in which all subsequent constructions take place.  
The interpretability of topological results therefore critically depends on this choice, and sensitivity analysis to alternative metrics is recommended.

\subsection{Step 3: Complex Construction}

From the distance matrix we construct a simplicial complex, a combinatorial object that approximates the data’s underlying shape.  
The most common choice is the \emph{Vietoris–Rips complex}, which includes all simplices (edges, triangles, tetrahedra, \emph{etc.}) whose vertices are pairwise within a given threshold~$\varepsilon$.  
For small~$\varepsilon$, the complex consists of isolated points; as~$\varepsilon$ grows, higher-dimensional simplices appear and clusters merge.

\smallbreak 
Alternative constructions include:
\begin{itemize}[leftmargin=1.2em]
  \item \textbf{\v{C}ech complex:} based on overlapping balls of radius~$\varepsilon/2$; topologically faithful but computationally heavier;
  \item \textbf{Witness complex:} uses a small set of landmarks to approximate topology for large datasets;
  \item \textbf{Alpha complex:} suitable when Euclidean coordinates are known; closely related to Delaunay triangulations.
\end{itemize}

Each of these complexes defines the same homology in the limit of small scales but may differ in computational cost and approximation accuracy. In business analytics, Rips complexes are typically preferred because they require only pairwise distances and scale well to high dimensions.

\subsection{Step 4: Persistence Computation}

Persistent homology tracks how the topology of the complex changes as the threshold~$\varepsilon$ varies.  
In computational practice, one constructs a sequence of complexes
\[
K_{\varepsilon_1} \subseteq K_{\varepsilon_2} \subseteq \cdots \subseteq K_{\varepsilon_m},
\]
known as a \emph{filtration}.  
Algorithms compute when each feature (connected component, loop, void) is born and when it dies as the scale increases \cite{ZomorodianCarlsson2005}.

The result is a barcode or persistence diagram summarizing the lifetime of features in different dimensions:
\begin{itemize}[leftmargin=1.2em]
  \item $H_0$: connected components (clusters or regimes);
  \item $H_1$: loops or cycles (periodicity, feedback);
  \item $H_2$: voids (high-dimensional diversification gaps).
\end{itemize}
Software such as \texttt{Ripser} (\cite{Tralie2018}) or \texttt{giotto-tda} (\cite{Giotto2021}) implements these computations efficiently and outputs visualizations directly.

\subsection{Step 5: Feature Extraction and Visualization}

Persistence summaries can be interpreted visually or transformed into numerical features for further analysis.

\paragraph{Barcodes and diagrams.}
Analysts can visually inspect bar lengths to identify persistent structure: long bars indicate stable, significant features; short bars correspond to noise.

\paragraph{Persistence images and landscapes.}
By mapping diagrams into grids or functional representations, one obtains fixed-size vectors suitable for machine-learning models \cite{Adams2017}.  
These representations have been used for regime classification, anomaly detection and sentiment analysis. 

\paragraph{Derived indicators.}
Aggregating lifetimes or computing their variance yields simple indices such as the \emph{Total Persistence} (overall structural complexity) or the proposed \emph{Topological Stability Index} (see Section~\ref{sec:theoretical}).

Visualization plays a central role in communicating results to non-technical audiences.  
Persistence diagrams, when presented alongside traditional plots (correlation matrices, PCA maps), reveal geometric aspects invisible to linear projections.

\subsection{Step 6: Downstream Analysis and Decision Support}

As mentioned earlier, TDA rarely stands alone; it complements existing analytical workflows.  
Once persistence summaries are computed, they can be incorporated into familiar frameworks:
\begin{itemize}[leftmargin=1.2em]
  \item \textbf{Clustering and segmentation:} use distances between persistence diagrams to cluster similar patterns;
  \item \textbf{Classification and forecasting:} feed persistence images into machine-learning models as geometric features;
  \item \textbf{Portfolio or network analysis:} monitor topological indicators as early-warning signals of regime shifts or structural stress;
  \item \textbf{Visualization dashboards:} combine barcodes with traditional KPIs for managerial interpretation.
\end{itemize}

Because persistence is scale-invariant and noise-resistant, these downstream tasks inherit its robustness, making decisions less sensitive to arbitrary parameter choices or data perturbations.

\subsection{Reporting and Reproducibility Checklist}

For reproducibility and comparability, each TDA study should report:
\begin{itemize}[leftmargin=1.2em]
  \item[-] the data preprocessing steps and windowing scheme used;
  \item[-] the distance metric and its parameters;
  \item[-] the type of complex and range of $\varepsilon$ values;
  \item[-] the homology dimensions analyzed ($H_0$, $H_1$, etc.);
  \item[-] software versions and computational settings.
\end{itemize}

Including such a checklist ensures that results can be validated and reused by other researchers and practitioners.

\medskip
\noindent In summary, the TDA pipeline provides a transparent, domain-independent sequence of geometric transformations from data to insight.
Each step adds interpretive meaning, transforming raw numerical data into a map of structural organization. The next section demonstrates this process through comparative case studies in consumer dynamics, equity markets and foreign exchange.

\section{Case Studies}\label{sec:case}

To illustrate how Topological Data Analysis (TDA) operates in practice and what kinds of insights it yields beyond traditional tools, 
we present three detailed case studies drawn from recent research.  
Each study applies the full TDA pipeline, from preprocessing and metric selection to persistence computation and interpretation, 
to a different analytical context: stock-market dynamics, consumer attention behavior, and currency co-movements.  
Together they demonstrate how topology complements classical and symbolic approaches, uncovering geometric structure, cyclic behavior and scale-dependent stability in complex economic systems.

\subsection{Stock Markets: Symbolic and Topological Patterns in European Equities}

\paragraph{Background and Motivation.}

Financial markets are dynamic, non-linear and strongly interconnected systems.  
The evolution of equities reflects overlapping sources of dependency, sectoral structure, macroeconomic conditions, and global shocks, that unfold at different time scales.  
Traditional linear tools, such as correlation matrices or principal component analysis (PCA), assume stationarity and Gaussian dependence.  
While effective for broad co-movement analysis, they often fail to capture regime transitions or local synchronization.

Symbolic representations, such as the \emph{Symbolic Aggregate Approximation} (SAX), provide an alternative perspective by compressing continuous time series into symbolic strings that encode qualitative trends.  
They are computationally efficient and highlight recurring motifs but lose the geometric continuity of trajectories.  
In contrast, \emph{Topological Data Analysis} (TDA) preserves the full metric structure of the data and describes its organization across multiple scales of similarity, making it well suited to detect regime changes and sectoral dynamics.

The study of \cite{HobbelhagenDiamantis2024} investigated how symbolic and topological methods differ in identifying structural relationships among large-cap European equities.  
The central question was whether persistent homology could detect evolving patterns of connectivity more effectively than symbolic representations, and whether combining the two could yield complementary insights.

\paragraph{Data and Preprocessing.}

The dataset consisted of the daily closing prices of 60 European equities covering eleven economic sectors (A–K in NACE classification), over the period May~2023–May~2024.  
To ensure comparability, prices were converted to log-returns and z-score normalized per stock.  
Twelve non-overlapping monthly windows (each of approximately 21 trading days) were extracted, forming a sequence of temporal slices that capture evolving market structure.  
Each window was analyzed separately.

Two symbolic representations were generated:
\begin{enumerate}[label=(\roman*), leftmargin=2em]
  \item \textbf{SAX:} each normalized time series was divided into equal-length segments; each segment was assigned a symbol according to its mean z-normalized value;
  \item \textbf{eSAX:} an extension encoding both minimum and maximum values per segment, preserving volatility information in addition to trend.
\end{enumerate}

For the TDA approach, pairwise similarities between stocks were expressed through the correlation-based distance
\[
d_{ij} = \sqrt{2(1-\rho_{ij})},
\]
where $\rho_{ij}$ is the Pearson correlation between daily returns of assets $i$ and $j$.  
This metric captures co-movement while being invariant to linear scaling.  
The resulting $60\times60$ distance matrix was used to construct Vietoris-Rips complexes at multiple thresholds~$\varepsilon$.  
Embedding parameters for phase-space reconstruction were selected using the false-nearest-neighbor (FNN) criterion with embedding dimension $m=7$ and delay $\tau=3$, ensuring sufficient reconstruction of dynamical structure.

\paragraph{Symbolic Analysis.}

Symbolic strings were first transformed into numerical feature vectors using piecewise aggregate approximation (PAA).  
Hierarchical clustering and $k$-means algorithms were then applied to compare the resulting symbolic feature spaces.

\begin{figure}[H]\centering
\includegraphics[width=1\textwidth]{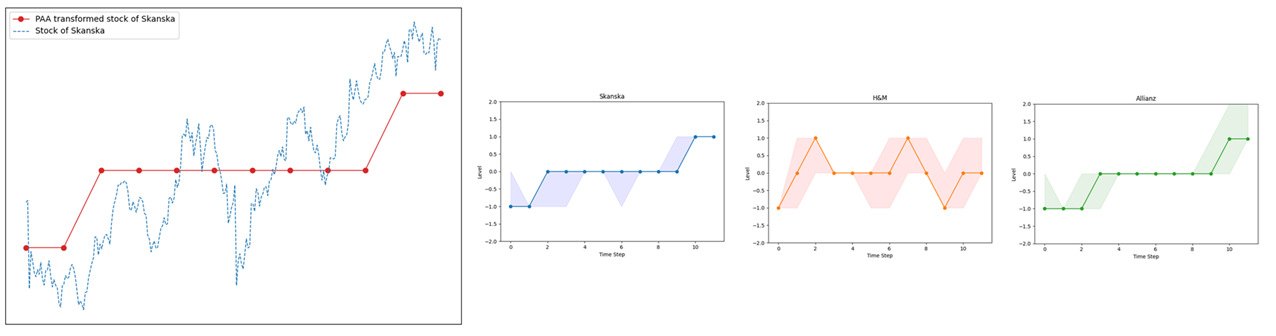}
\caption{Illustration of symbolic transformations.  
Left: Piecewise Aggregate Approximation (PAA) of stock prices.  
Right: SAX and extended SAX (eSAX) representations for Skanska, H\&M, and Allianz, showing enhanced sensitivity to volatility range.  
(Source: Hobbelhagen \& Diamantis, 2024.)}
\label{fig:sax_example}
\end{figure}

\begin{figure}[H]\centering
\includegraphics[width=1\textwidth]{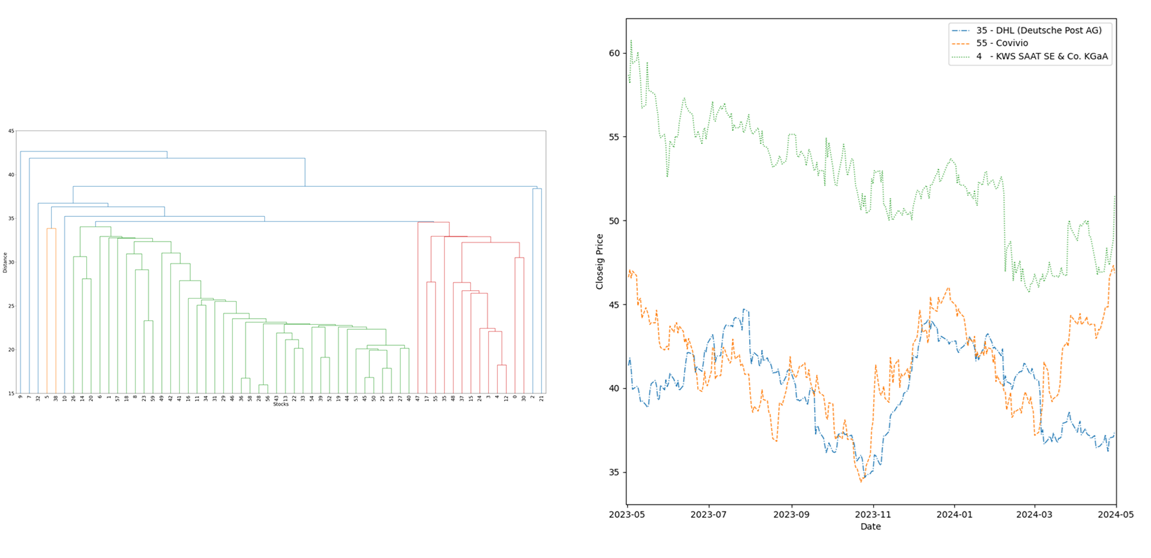}
\caption{Hierarchical clustering of symbolic representations.  
Left: SAX-based dendrogram showing broad macro-grouping across sectors.  
Right: three sample equities from the same cluster illustrating diverging short-term behavior despite symbolic similarity.  
(Source: Hobbelhagen \& Diamantis, 2024.)}
\label{fig:sax_clusters}
\end{figure}

Symbolic clustering revealed general patterns of market behavior.  
Figure~\ref{fig:sax_clusters} displays a hierarchical clustering dendrogram built from SAX features.  
During low-volatility periods, most equities formed a single macro-cluster, while during turbulent intervals multiple micro-clusters emerged.  
However, sectors with distinct volatility profiles, such as utilities and healthcare versus cyclicals, were occasionally grouped together when their mean trajectories coincided.

\paragraph{Topological Analysis.}

\begin{figure}[H]\centering
\includegraphics[width=0.9\textwidth]{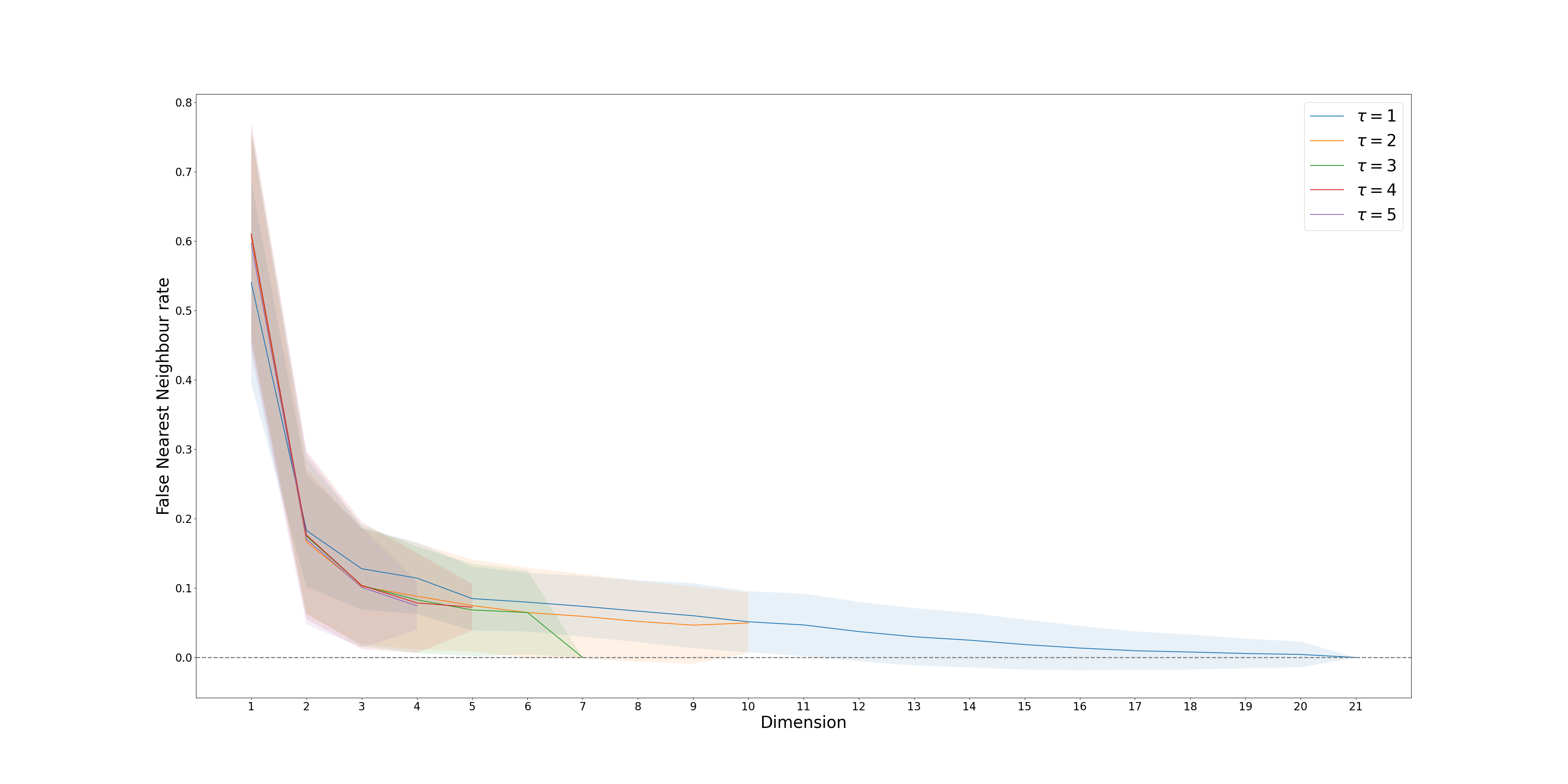}
\caption{Average false-nearest-neighbor (FNN) ratios across embedding dimensions $m$ and delays~$\tau$.  
The chosen parameters $m=7$, $\tau=3$ minimize false neighbors, ensuring reliable reconstruction of temporal structure.  
(Source: Hobbelhagen \& Diamantis, 2024.)}
\label{fig:fnn}
\end{figure}

To explore finer-scale geometric dependencies, TDA was applied to the same windows.  
For each distance matrix, persistent homology was computed in dimensions $H_0$ and $H_1$, and persistence diagrams and landscapes were derived.  
The FNN analysis guiding embedding selection is summarized in Figure~\ref{fig:fnn}.

\begin{figure}[H]\centering
\includegraphics[width=1\textwidth]{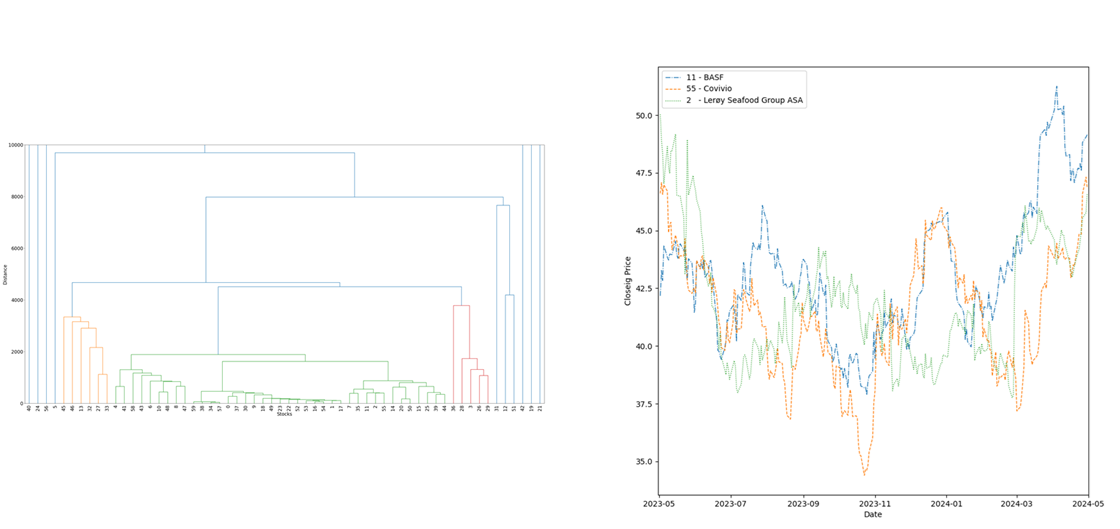}
\caption{Hierarchical clustering in the TDA feature space.  
Left: dendrogram based on persistence features ($H_0$, $H_1$).  
Right: three example equities from the same TDA cluster showing consistent local geometry and correlated short-term dynamics.  
(Source: Hobbelhagen \& Diamantis, 2024.)}
\label{fig:tda_clusters}
\end{figure}

Persistence diagrams were converted into summary statistics, total persistence, lifetime variance, and number of long bars, and used as geometric features for clustering and temporal comparison.  
The TDA dendrogram (Figure~\ref{fig:tda_clusters}) shows clearer separation between clusters and isolates outliers more distinctly than the symbolic counterpart, indicating stronger structural differentiation in the correlation geometry.

\paragraph{Results and Comparative Insights.}

\begin{figure}[H]\centering
\includegraphics[width=0.72\textwidth]{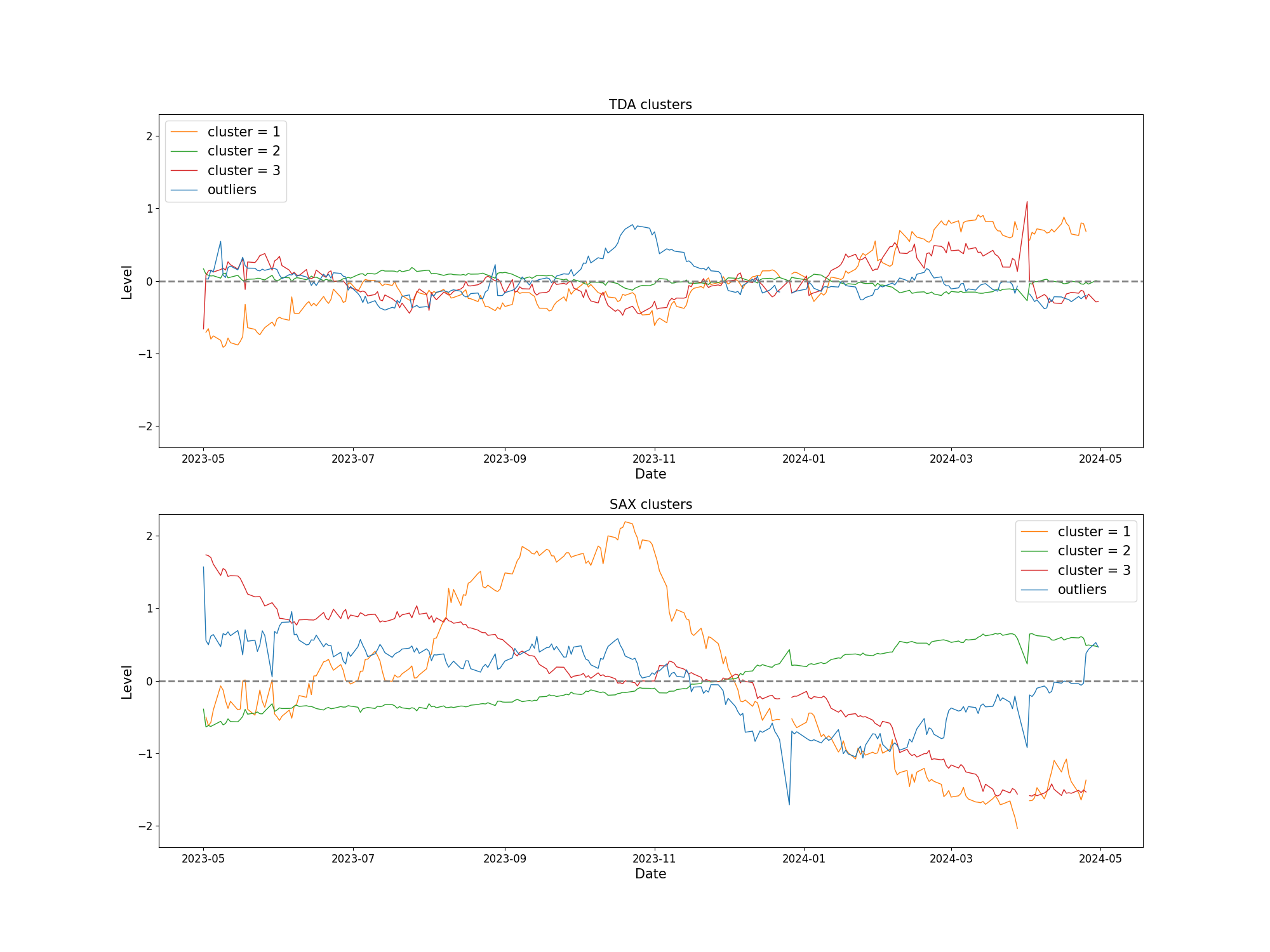}
\caption{Average normalized stock prices per cluster.  
Top: clusters obtained via TDA; bottom: clusters from SAX.  
TDA preserves local geometric detail and separates sectors with distinct short-term dynamics.  
(Source: Hobbelhagen \& Diamantis, 2024.)}
\label{fig:tda_sax_comparison}
\end{figure}

Figure~\ref{fig:tda_sax_comparison} compares the average normalized price trajectories of clusters obtained via SAX and TDA.  
Both methods reveal macro-level coherence but differ in their treatment of micro-level fluctuations.  
SAX merges assets with broadly similar trends even if their intra-window dynamics diverge, while TDA distinguishes stocks with subtle but persistent differences in temporal synchronization.

\begin{figure}[H]\centering
\includegraphics[width=0.8\textwidth]{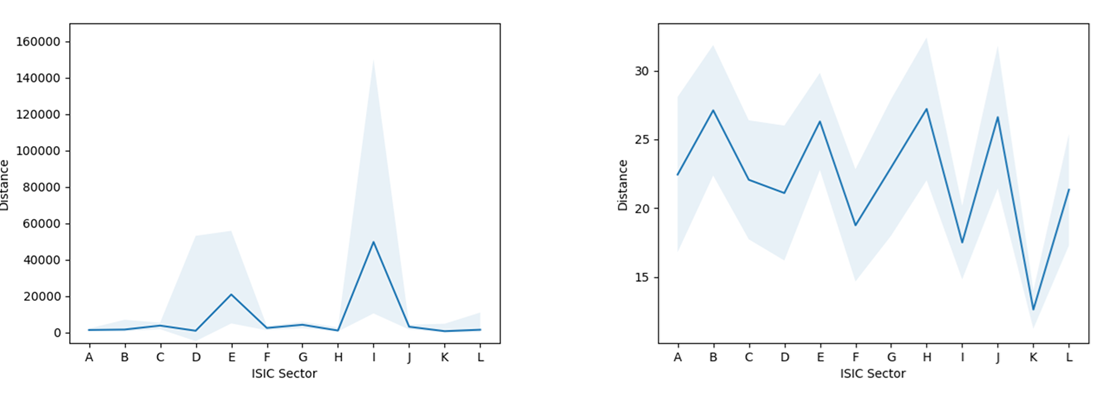}
\caption{Within-sector median distances for TDA (left) and SAX (right).  
Lines show means; shaded regions indicate standard deviations across time windows.  
TDA reveals stronger temporal fluctuations in sectoral cohesion, especially in sectors D, E, and I.  
(Source: Hobbelhagen \& Diamantis, 2024.)}
\label{fig:sector_dispersion}
\end{figure}

Sector-level analysis revealed that both methods grouped similar industries (construction and finance), but TDA highlighted higher dispersion in others (sectors D, E, and I), pointing to greater internal heterogeneity.  
Figure~\ref{fig:sector_dispersion} displays within-sector median distances for TDA and SAX, averaged across windows.  
The shaded areas correspond to standard deviations, visualizing temporal variability in cohesion.

\begin{figure}[H]\centering
\includegraphics[width=0.55\textwidth]{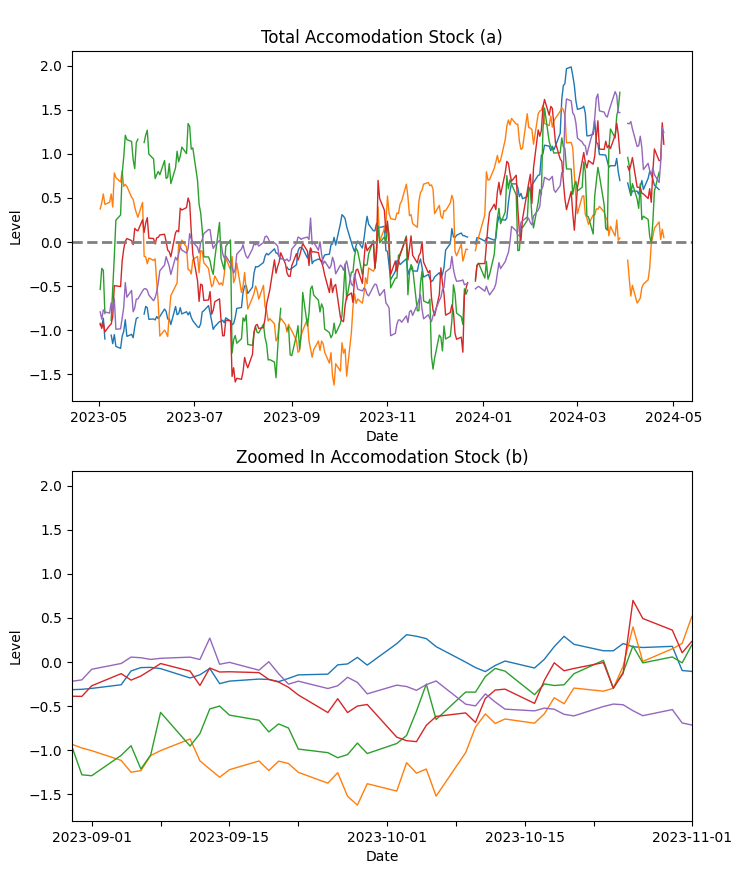}
\caption{Stock movements in the accommodation and food service sector (I).  
(a) overall trend; (b) zoomed-in local fluctuations.  
TDA captures amplitude-level structure that SAX smooths out.  
(Source: Hobbelhagen \& Diamantis, 2024.)}
\label{fig:sector_example}
\end{figure}

A closer inspection of the accommodation and food service sector (I) demonstrates how symbolic and topological methods diverge in sensitivity to local variation.  
SAX captures general directionality, whereas TDA distinguishes subtle structural differences in shape and local amplitude (Figure~\ref{fig:sector_example}).

\paragraph{Discussion and Economic Interpretation.}

The comparative analysis demonstrates that TDA captures structural features of market dynamics invisible to symbolic or linear models.  
While SAX efficiently summarizes broad market regimes, its discretization loses sensitivity to intermediate-scale variability.  
TDA, by retaining continuous geometric information, identifies scale-dependent transitions: short-lived features correspond to transient fluctuations, whereas long-lived topological components and loops represent persistent co-movement structures.

Economically, persistent $H_0$ features correspond to stable sectoral clusters of equities that move together across thresholds, whereas persistent $H_1$ loops describe cyclical or alternating relationships between sectors.  
High variance in lifetimes, interpretable as \emph{topological volatility}, acts as a geometric stress indicator, often rising before spikes in overall market volatility.

\paragraph{Managerial Implications.}

For portfolio managers, monitoring persistent topological features offers complementary information to classical risk measures.  
Persistent $H_1$ loops reveal latent cyclic dependencies among sectors that covariance-based models may overlook.  
Tracking the birth and death of these features over time enables early identification of diversification breakdowns or regime transitions.  
Because TDA is scale-invariant and noise-resistant, it integrates seamlessly into risk dashboards without heavy parametric calibration.

In summary, topology enriches symbolic time-series analysis by introducing a continuous, scale-aware perspective on structure.  
While SAX and eSAX efficiently encode dominant trends, TDA uncovers hidden organization in market geometry, supporting a more robust understanding of financial interconnectedness.

\subsection{Consumer Attention Dynamics: The Shape of Online Behavior}

\paragraph{Background and Motivation.}

Consumer attention is among the most volatile and informative signals in modern digital markets.  
Online interest rises and fades rapidly as global events, technologies and cultural trends compete for visibility.  
Traditional analytics, such as moving averages, correlations, autoregressive or seasonal-trend models, capture temporal dependence but overlook the \emph{geometry} of behavioral change.  
Similarity at a fixed scale may group together categories that share seasonality but differ in how these patterns evolve through time.

The study of \cite{BeretaDiamantis2025} investigated whether the \emph{shape} of attention trajectories, rather than only their amplitude or correlation, could uncover latent behavioral regimes.  
By comparing symbolic (SAX/eSAX) and topological representations of Google Trends time series, the paper introduced TDA as a novel analytical lens for marketing science and behavioral economics.

\paragraph{Data and Preprocessing.}

Weekly Google Trends indices were collected for 20 consumer-related keywords covering six thematic categories:
\begin{enumerate}[leftmargin=2em,label=(\roman*)]
  \item \emph{Technology and AI} (e.g., ``ChatGPT'', ``machine learning'');
  \item \emph{Digital Work and Education};
  \item \emph{Finance and Economy};
  \item \emph{Sustainability and Environment};
  \item \emph{Personal Devices};
  \item \emph{Common Products}.
\end{enumerate}
The observation period spanned 2019--2024.  
Values were normalized to $[0,1]$ for cross-term comparability and converted into overlapping 12-week windows with a one-week stride, generating about 250 local segments per keyword (points in~$\mathbb{R}^{12}$).  
Each segment was z-normalized to remove long-term drift, producing a manifold of attention trajectories.

\paragraph{Methodology.}

Two complementary representations were compared:

\begin{itemize}[leftmargin=1.5em]
  \item \textbf{Symbolic representation:} normalized segments were transformed via SAX and eSAX into symbolic words.  
  Distances between windows were computed as Euclidean distances between their Piecewise Aggregate Approximation (PAA) vectors with alphabet size $a = 7$.
  \item \textbf{Topological representation:} the same windows were embedded in a metric space using correlation-based distance  
  $d_{ij}=\sqrt{2(1-\rho_{ij})}$, where $\rho_{ij}$ is the Pearson correlation between windows.  
  Persistent homology was then computed on Vietoris–Rips complexes for increasing thresholds~$\varepsilon$, up to homology dimension~$H_1$.
\end{itemize}

Persistence diagrams were transformed into persistence landscapes for interpretability and aggregated over time to track the evolution of long-lived structures in attention dynamics.

\begin{figure}[H]\centering
\includegraphics[width=1\textwidth]{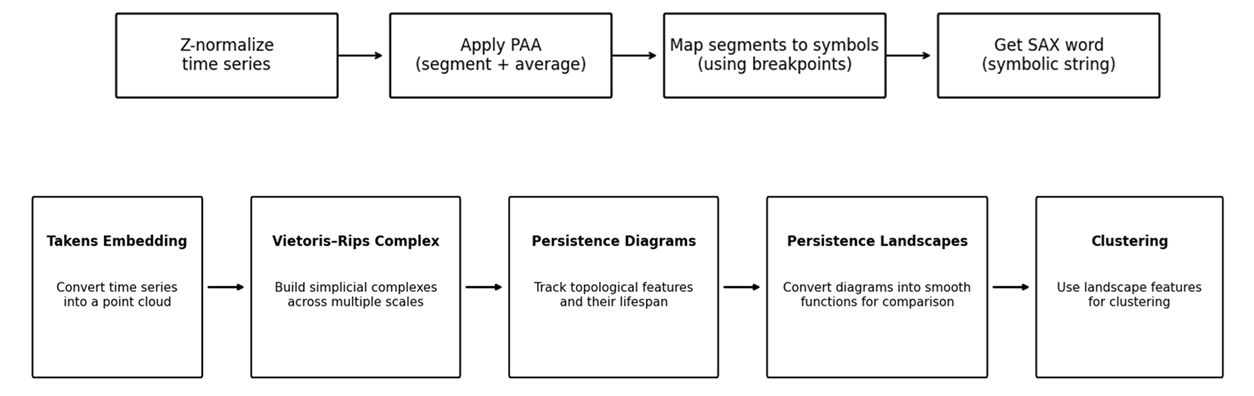}
\caption{Analytical workflow for symbolic (SAX/eSAX) and topological (TDA) representations.  
Top: SAX pipeline from normalization to symbolic distance matrix.  
Bottom: TDA pipeline from correlation distance to persistence diagrams.  
(Source: Bereta \& Diamantis, 2025.)}
\label{fig:consumer-pipeline}
\end{figure}

\paragraph{Results.}

Symbolic methods reproduced well-known seasonal cycles, for example, winter peaks in ``holiday shopping'' and early-year increases in ``budgeting'' and ``taxes''.  
However, SAX tended to merge keywords with similar periodicities but different volatility structures:  
``AI'' and ``blockchain'' shared yearly rhythms yet diverged sharply during hype bursts, which symbolic clustering masked.

\begin{figure}\centering
\includegraphics[width=1\textwidth]{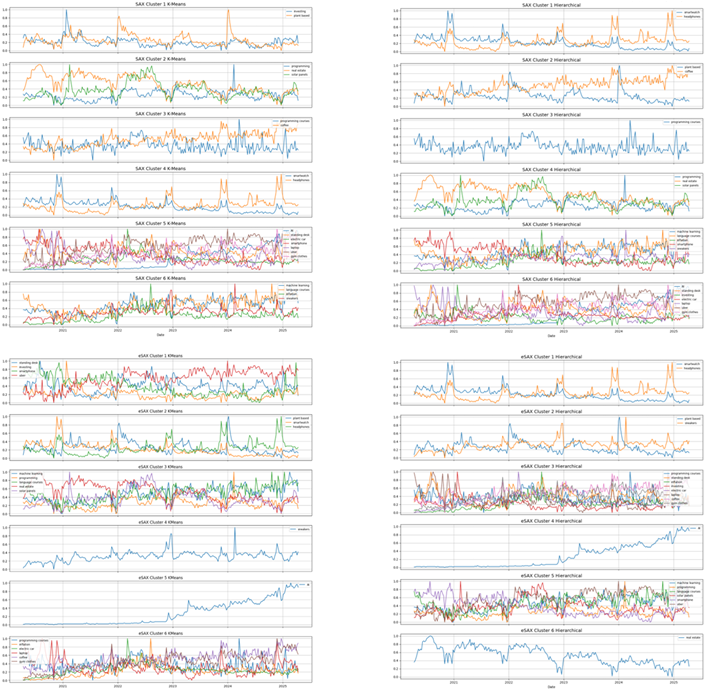}
\caption{Symbolic clustering of 20 Google Trends keywords (2019--2024) based on SAX and eSAX representations.  
Top: $k$-means clustering for (left) SAX and (right) eSAX feature spaces.  
Bottom: hierarchical clustering for the same representations.  
Both methods capture broad seasonal regularities but tend to merge keywords with similar periodicity while overlooking differences in volatility and hype intensity.  
(Source: Bereta \& Diamantis, 2025.)}
\label{fig:consumer-clusters}
\end{figure}

TDA disentangled these subtleties by examining the geometry of trajectories.  
Persistent $H_0$ features revealed coherent behavioral regimes: stable attention clusters within domains such as sustainability or finance.  
Persistent $H_1$ features captured loops in collective attention, where topics moved through phases of excitement, decline and renewed interest, mirroring the rhythm of hype and stabilization cycles.

\begin{figure}[H]\centering
\includegraphics[width=0.95\textwidth]{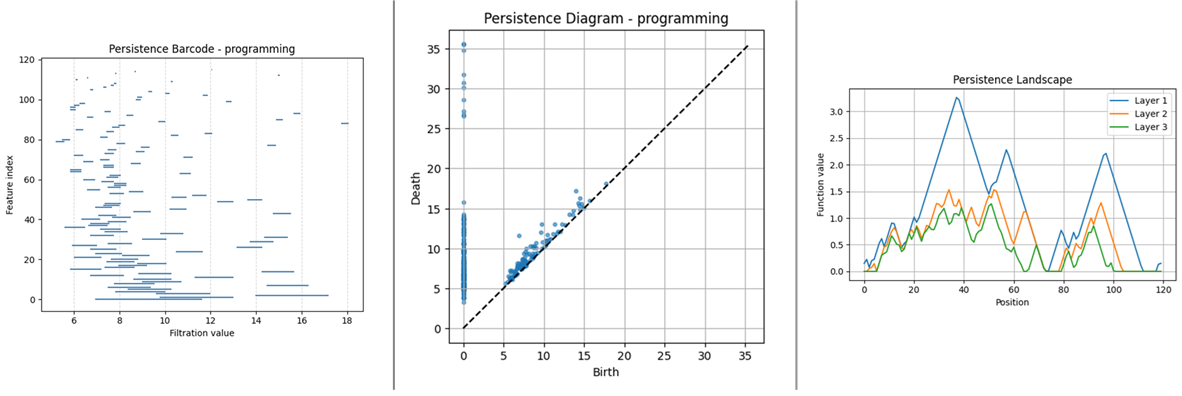}
\caption{
Complementary representations of persistent homology for the keyword ``Programming''.  
Left: persistence barcode showing feature lifetimes across scales.  
Middle: persistence diagram plotting birth–death coordinates, where points farther from the diagonal represent long-lived, stable structures.  
Right: persistence landscape providing a functional summary of persistence, useful for statistical comparison and feature extraction.  
Together these views illustrate how TDA captures both the topology and the stability of consumer attention dynamics.  
(Source: Bereta \& Diamantis, 2025.)
}
\label{fig:consumer-programming}
\end{figure}

Specific examples illustrate the interpretive power of topology:
\begin{itemize}[leftmargin=1.5em]
  \item \emph{Technology and AI:} long-lived $H_1$ loops captured repeated surges of attention (e.g., the launch of ChatGPT and subsequent normalization).  
  \item \emph{Digital Work:} persistent cycles mirrored the pandemic’s work-pattern transitions (lockdown $\rightarrow$ hybrid $\rightarrow$ normalization).  
  \item \emph{Finance and Economy:} reductions in $H_0$ persistence around inflation spikes indicated fragmentation of public focus.
\end{itemize}

\begin{figure}[H]\centering
\includegraphics[width=1\textwidth]{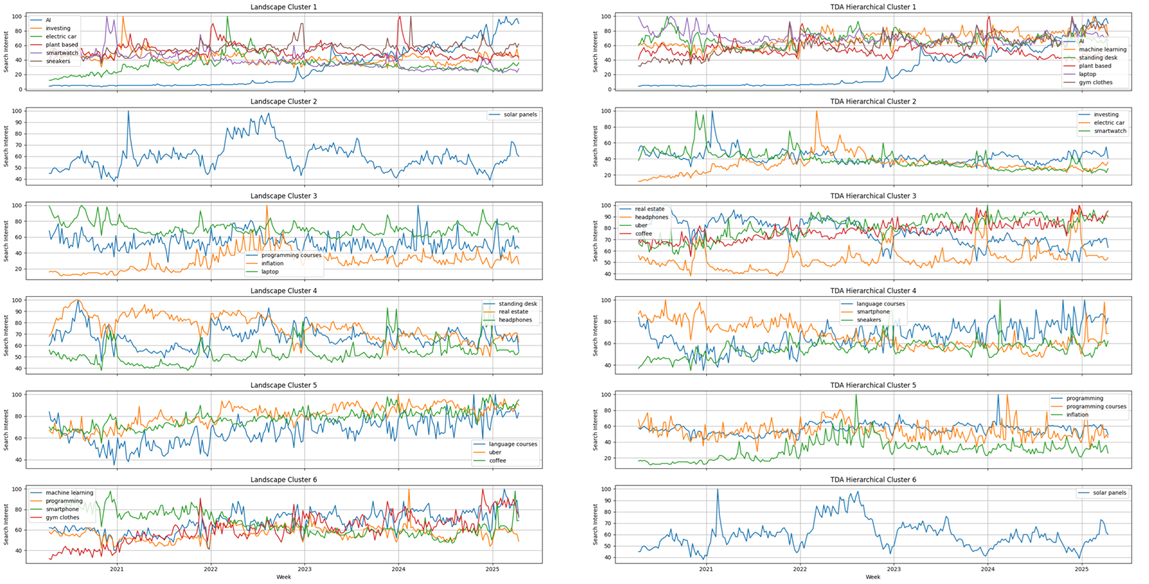}
\caption{
Clustering of 20 Google Trends keywords in the topological feature space.  
Left: $k$-means clustering using TDA-derived features.  
Right: hierarchical clustering on the same feature set.  
Compared to symbolic methods, TDA-based clusters exhibit greater compactness and clearer thematic separation, confirming the added geometric discriminability of persistent features.  
(Source: Bereta \& Diamantis, 2025.)
}
\label{fig:consumer-tda-clusters}
\end{figure}

To visualize these quantitative improvements, Figure~\ref{fig:consumer-tda-clusters} presents the clustering results obtained directly in the TDA feature space.  
Compared with the symbolic clusters shown earlier, the topological representation yields sharper boundaries and thematically coherent groups, confirming the discriminative power of persistent features.

Quantitatively, TDA outperformed symbolic approaches in internal clustering validity:  
Silhouette score 0.61 vs.\ 0.44 for SAX, and Davies–Bouldin index 1.15 vs.\ 1.86.  
The total persistence of $H_1$ features correlated strongly ($r=0.72$) with volatility in weekly Google Trends values, confirming that topological variability tracks bursts of collective attention.

\begin{table}[h!]
\centering
\caption{Clustering performance metrics comparing symbolic and topological feature spaces.  
Higher Silhouette and lower DBI values indicate clearer separation.  
(Source: Bereta \& Diamantis, 2025.)}
\label{tab:final-model-comparison}
\begin{tabular}{C{2.2cm} L{4.2cm} C{2.2cm} C{2.2cm} C{2.2cm}}
\toprule
& \textbf{Metric} & \textbf{SAX} & \textbf{eSAX} & \textbf{TDA} \\
\midrule
& Silhouette Score         & 0.320 & 0.218 & 0.146 \\
K-Means & &&&\\
& Davies–Bouldin Index     & 0.749 & 0.777 & 1.133 \\
\midrule
& Silhouette Score   & 0.355 & 0.308 & \textbf{0.375} \\
Hierarchical & &&&\\
& Davies–Bouldin Index & {\bf 0.618} & 0.652 & 0.723 \\
\bottomrule
\end{tabular}
\end{table}

\paragraph{Interpretation and Discussion.}

TDA adds a geometric dimension to behavioral analytics.  
Where SAX compresses data into symbolic motifs, persistent homology links local variations across scales, revealing when attention cycles re-emerge or decay.  
In marketing terms:
\begin{itemize}[leftmargin=1.5em]
  \item persistent $H_0$ bars $\Rightarrow$ stable consumer interests or categories with long-term relevance;
  \item persistent $H_1$ bars $\Rightarrow$ cyclical or event-driven attention loops (recurrent hype);
  \item short bars $\Rightarrow$ transient curiosity or viral anomalies.
\end{itemize}

These interpretations distinguish \emph{structural interest} (persistent components) from \emph{episodic attention} (short loops).  
Monitoring persistence lifetimes thus quantifies \emph{attention durability}, namely, the degree to which engagement withstands trend noise.

\paragraph{Managerial Implications.}

For firms, persistence provides actionable intelligence:
\begin{itemize}[leftmargin=1.5em]
  \item allocate resources to categories with long-lived $H_0$ features (steady demand);
  \item synchronize marketing with phases of emerging $H_1$ loops (resurging hype);
  \item detect consumer fatigue when total persistence declines across themes.
\end{itemize}
For instance, sustainability maintained consistently high $H_0$ lifetimes, signaling deep and stable concern, whereas technology and AI showed high $H_1$ volatility typical of innovation-driven cycles.

\paragraph{Broader Significance.}

Beyond empirical findings, the study demonstrates the versatility of TDA in behavioral analytics.  
It unifies symbolic summarization and geometric analysis within a single pipeline, revealing both macro- and micro-dynamics of attention.

\subsection{Currency Networks: Persistent Homology for FX Co-movements}

\paragraph{Background and Motivation.}

Foreign exchange (FX) markets form one of the largest and most interconnected financial systems in the world.  
Currencies co-move through trade relations, interest-rate differentials, and capital flows that evolve over time.  
Traditional econometric analyses rely on correlations or co-integration to study these dependencies, assuming linearity and stationarity.  
However, exchange-rate dynamics often display nonlinear and multi-scale behavior, with structural breaks and regime transitions that such tools cannot fully capture.

The study in \cite{DeFavereauDiamantis2025} investigated whether \emph{persistent homology} can provide a robust, interpretable description of structural organization within FX markets.  
By comparing topological representations of currency dynamics with classical and statistical methods, the paper demonstrated that TDA can uncover meaningful co-movement patterns that remain stable across scales.

\paragraph{Data and Preprocessing.}

The dataset consisted of monthly logarithmic returns of 13 major currencies relative to the euro:
AUD, BRL, CHF, CNY, GBP, INR, JPY, KRW, RUB, THB, TRY, USD, and ZAR.  
Two datasets were analyzed: one covering 2000–2022 (including RUB) and one extended to 2024 (excluding RUB).  
All series were standardized (z-scores) to ensure comparability across currencies.

Each currency was analyzed individually through a delay embedding with window length $d=4$ and delay $\tau=1$, transforming each time series into a point cloud in $\mathbb{R}^4$.  
This embedding preserves local temporal geometry and allows persistent homology to detect recurring structural patterns in returns.

\begin{figure}[H]\centering
\includegraphics[width=0.46\textwidth]{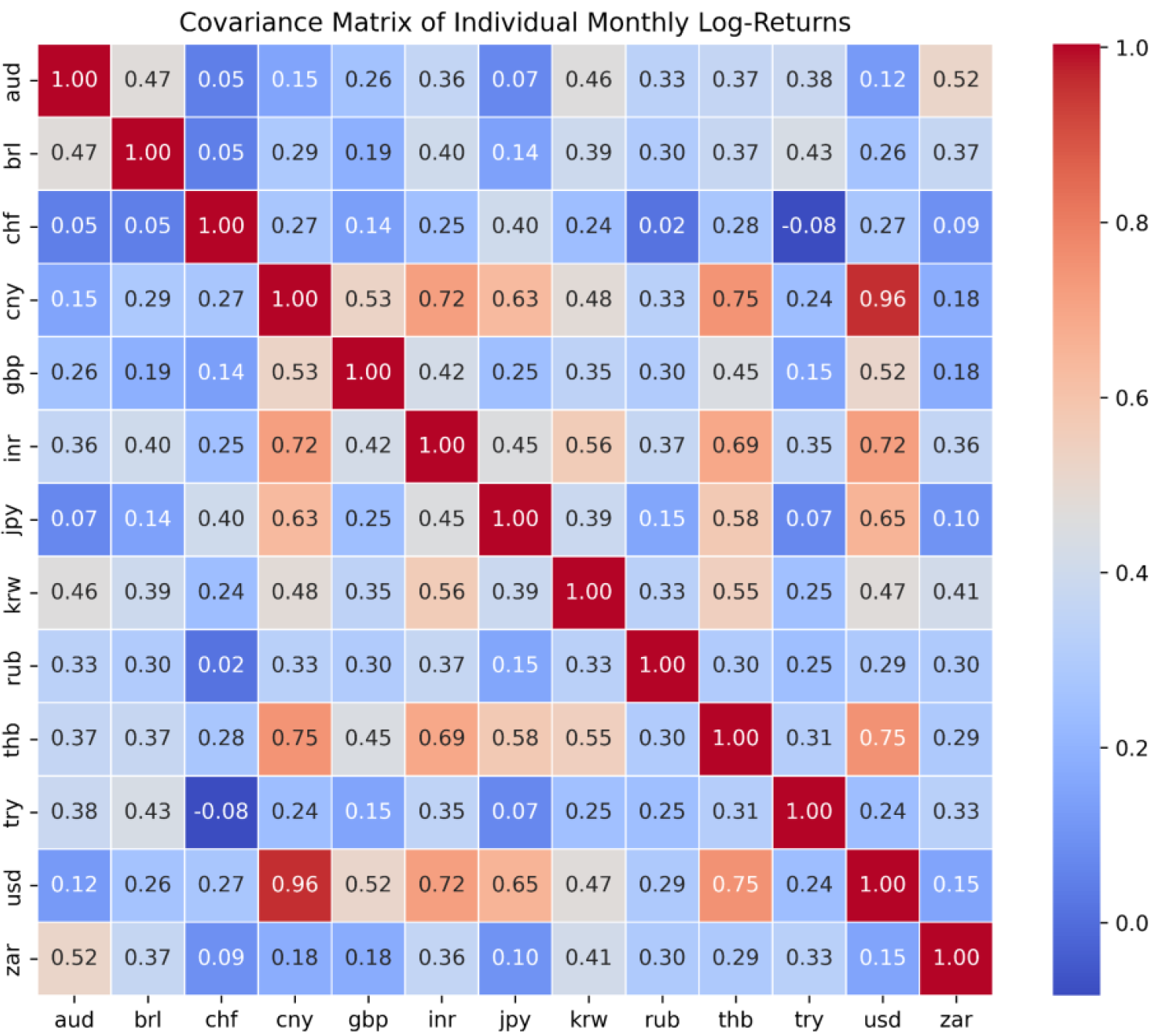}\hfill
\includegraphics[width=0.46\textwidth]{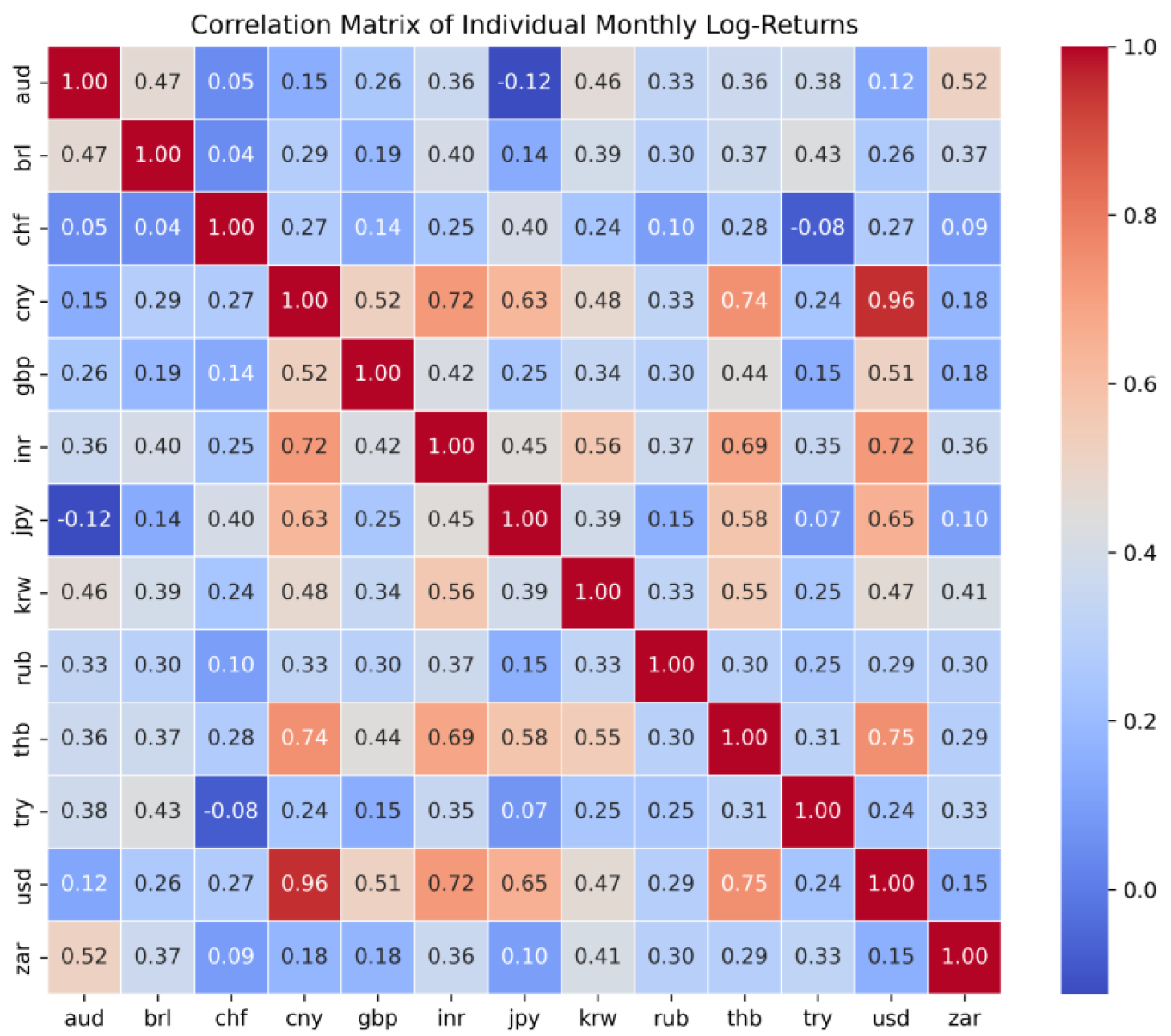}
\caption{Statistical dependence structure of standardized monthly FX returns.  
Left: covariance heatmap. Right: cross-correlation heatmap (maximum lag = 1 month).  
(Source: De Favereau de Jeneret \& Diamantis, 2025.)}
\label{fig:fx-stat-heatmaps}
\end{figure}

\paragraph{Methodology.}

Three approaches were compared:

\begin{enumerate}[label=(\roman*),leftmargin=2em]
  \item \textbf{Classical approach:} hierarchical and $k$-means clustering on statistical features derived from the standardized return series and their correlation matrices;
  \item \textbf{Spectral approach:} principal component analysis (PCA) on the covariance matrix to extract dominant co-movement modes;
  \item \textbf{Topological approach:} computation of persistent homology on the delay-embedded point clouds.  
  Vietoris–Rips filtrations were applied to obtain persistence diagrams in $H_0$ and $H_1$,  
  and pairwise dissimilarities between currencies were quantified via Wasserstein distances between diagrams.  
  For $k$-means clustering, the Wasserstein matrix was embedded in Euclidean space through multidimensional scaling (MDS), while hierarchical clustering operated directly on the Wasserstein distances.
\end{enumerate}

\begin{figure}[H]\centering
\includegraphics[width=0.55\textwidth]{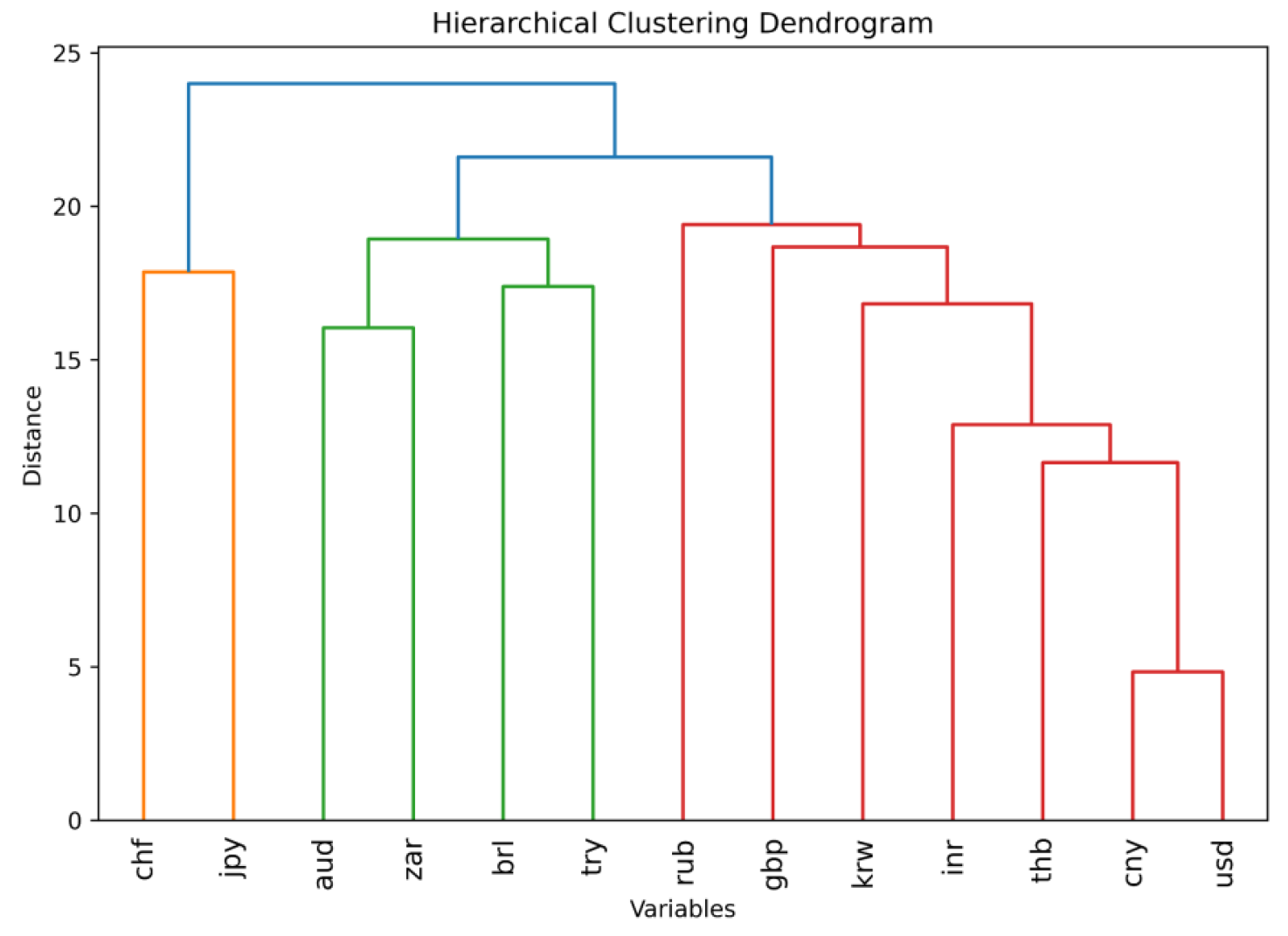}
\caption{Hierarchical clustering on statistical features.  
(Source: De Favereau de Jeneret \& Diamantis, 2025.)}
\label{fig:fx-stat-hclust}
\end{figure}

\paragraph{Results.}

\begin{figure}[H]\centering
\includegraphics[width=1\textwidth]{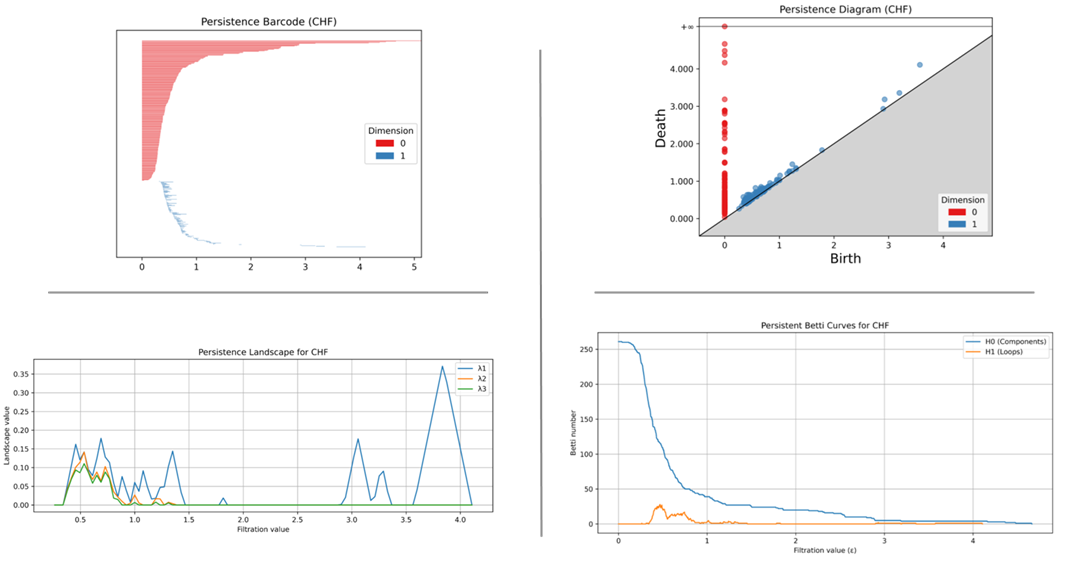}
\caption{Topological summaries for the CHF currency.  
Persistence barcode and diagram (top) and their functional analogues: persistence landscape and Betti curve (bottom).  
(Source: De Favereau de Jeneret \& Diamantis, 2025.)}
\label{fig:fx-tda-examples}
\end{figure}

Classical correlation-based clustering recovered broad regional groupings but failed to capture nonlinear distinctions between currencies.  
The topological approach, in contrast, revealed richer structural information.  
Persistence diagrams and barcodes displayed long-lived $H_0$ features; stable regimes of behavior within specific currencies, and occasional $H_1$ features indicating cycles in their embedded dynamics.  
Illustrative examples for CHF, GBP and THB showed clear differences in geometric complexity. In Figure~\ref{fig:fx-tda-examples} we illustrate these features for the CHF currency.

Quantitatively, the TDA-based models outperformed statistical baselines on internal-validity metrics.  
Average Silhouette and Calinski–Harabasz indices were higher for both $k$-means and hierarchical clustering in the topological space,  
indicating more compact and better-separated clusters.  
Specifically, the Statistical $k$-means (0.110, 2.657) and Statistical hierarchical (0.111, 2.942) models were surpassed by TDA $k$-means (0.191, 4.850) and TDA hierarchical (0.182, 5.905).

\begin{figure}[H]\centering
\includegraphics[width=0.61\textwidth]{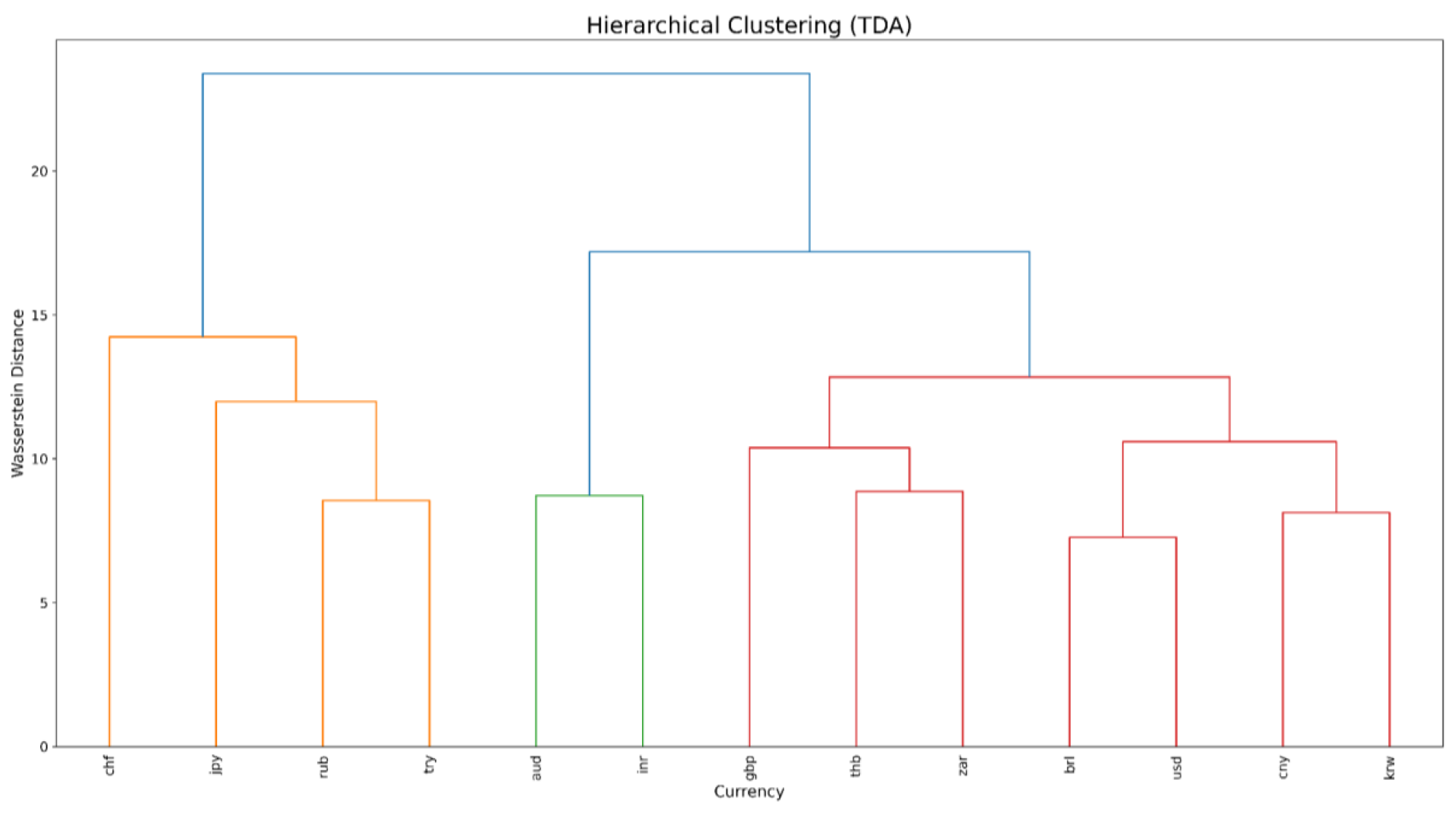}
\caption{Hierarchical clustering of 13 major currencies using persistence-based features.  
(Source: De Favereau de Jeneret \& Diamantis, 2025.)}
\label{fig:fx-tda-clusters}
\end{figure}

\paragraph{Interpretation and Discussion.}

Topology captures not only the strength but the \emph{structure} of currency relationships.  
Long-lived $H_0$ features correspond to persistent behavioral regimes, that is, currencies that maintain coherent dynamics over time, while the emergence of $H_1$ features reflects oscillatory or cyclical tendencies within the embedded trajectories.  
Because persistence diagrams are stable under small perturbations, these patterns represent genuine geometric organization rather than statistical noise.

\paragraph{Managerial and Policy Implications.}

For portfolio managers and policymakers, TDA offers a new diagnostic dimension.  
It provides a scale-independent and noise-resistant measure of similarity that complements correlation analysis.  
Stable topological components identify robust behavioral regimes, while increased short-lived topological features indicate potential instability or transitions.  
Combining both perspectives yields a more complete picture of systemic structure and market interdependence.

\begin{table}[H]\centering
\caption{TDA-derived $k$-means clusters (Wasserstein + MDS embedding).}
\label{tab:fx-tda-kmeans}
\begin{tabular}{ll}
\toprule
Cluster & Currencies \\
\midrule
1 & GBP, INR \\
2 & AUD, BRL, CNY, JPY, KRW, THB, TRY, USD, ZAR \\
3 & CHF, RUB \\
\bottomrule
\end{tabular}
\end{table}

\begin{table}[H]\centering
\caption{Clustering performance metrics across models (average Silhouette; Calinski–Harabasz).}
\label{tab:fx-metrics}
\begin{tabular}{lcc}
\toprule
Model & Silhouette & CH \\
\midrule
Statistical $k$-means & 0.110 & 2.657 \\
Statistical hierarchical & 0.111 & 2.942 \\
TDA $k$-means & 0.191 & 4.850 \\
TDA hierarchical & 0.182 & 5.905 \\
\bottomrule
\end{tabular}
\end{table}

\paragraph{Broader Significance.}

This study demonstrates that persistent homology provides a compact, geometric description of global financial interdependence.  
By summarizing the \emph{shape} of co-movement rather than its magnitude, TDA reveals structural resilience and fragility in a transparent, interpretable manner.  
Such tools could enrich macro-prudential frameworks by translating complex dependency patterns into robust topological diagnostics for systemic-risk monitoring.

\subsection{Comparative Synthesis and Broader Implications}

The three case studies presented above, spanning equity markets, consumer attention and foreign-exchange dynamics, demonstrate the versatility of Topological Data Analysis (TDA) as a unifying framework for studying structure in complex economic systems.  
Despite differences in scale, domain and data type, a consistent pattern emerges: topology captures how relationships evolve across similarity thresholds, uncovering organization, cyclicity and transitions that symbolic or linear methods tend to smooth out.

\paragraph{Cross-domain regularities.}
At a high level, all three applications reveal a shared geometric logic:
\begin{itemize}[leftmargin=1.2em]
  \item \textbf{Components ($H_0$):} Persistent connected components correspond to stable regimes; market sectors, consumer clusters or regional currency blocs, that remain cohesive across varying similarity thresholds.
  \item \textbf{Loops ($H_1$):} Persistent cycles indicate recurrent feedback mechanisms, such as alternating investor sentiment, product-hype phases or hedging loops among currencies.
  \item \textbf{Instability:} In every domain, bursts in total persistence or lifetime variance coincide with transitions-volatility spikes in equities, attention shifts in consumers, and contagion episodes in FX markets.
\end{itemize}

These analogies suggest that persistence plays a universal role as a measure of \emph{structural stability}:  
long-lived features correspond to resilient organization, while short-lived ones capture transient or noisy behavior.

\paragraph{Symbolic vs.\ Topological perspectives.}
The studies also highlight the complementarity between symbolic and topological representations.  
Symbolic methods (SAX, eSAX) are efficient at compressing information and detecting recurring temporal motifs, offering a discrete language for interpreting behavioral and market patterns.  
However, they discretize continuous variation and may merge distinct regimes into coarse symbolic categories.  
TDA, by contrast, retains the geometric continuity of data, exposing nonlinear dependencies and gradual transitions between regimes.  
Symbolic features thus excel at \emph{pattern summarization}, whereas topological features excel at \emph{structure detection}.  
Together they define a hybrid paradigm for modern analytics, combining the interpretability of symbolic methods with the robustness and depth of topology.

\paragraph{Methodological convergence.}
Beyond their domain-specific results, all three studies share a reproducible and conceptually simple computational workflow:
\begin{enumerate}[leftmargin=2em]
  \item Define a domain-relevant distance (Euclidean, correlation or symbolic);
  \item Construct a simplicial complex (typically Vietoris–Rips);
  \item Compute persistent homology ($H_0$, $H_1$);
  \item Interpret long-lived features as structural invariants.
\end{enumerate}
This four-step process generalizes seamlessly across contexts, from high-frequency trading data to macroeconomic aggregates, showing that TDA is not bound to a particular application but to a principle: that structure emerges through persistence across scales.

\paragraph{Managerial and analytical insights.}
From a practical perspective, persistence-based indicators can inform decision-making in diverse analytical settings:
\begin{itemize}[leftmargin=1.2em]
  \item In \emph{finance}, topological persistence reflects changes in structural connectivity among assets, complementing volatility and correlation-based measures of market stress.  
  \item In \emph{marketing}, it quantifies the durability and cyclicity of consumer attention, distinguishing stable interest from transient hype.  
  \item In \emph{macroeconomics}, it visualizes the geometry of interdependence and diversification across currencies, regions, or sectors.
\end{itemize}
These insights are inherently multi-scale: topology imposes no fixed parameter but tracks how relational patterns endure or vanish as the similarity threshold changes, offering a richer and more interpretable view of systemic organization.

\paragraph{Toward theoretical unification.}
Conceptually, the three case studies converge on a central theme: data in modern business and finance are rarely linear, and their complexity cannot be captured by variance or correlation alone.  
TDA extends the analytical vocabulary by introducing geometric notions, i.e. connectivity, cycles and persistence, into quantitative reasoning.  
Persistence diagrams can therefore be viewed as \emph{structural fingerprints} of complex systems, providing interpretable and comparable representations across domains.

\paragraph{Looking forward.}
The unifying message of these case studies is that \emph{the shape of data carries meaning}.  
By shifting focus from variables to relationships, TDA allows analysts to visualize and quantify organization itself.  
The next section develops these insights theoretically, examining how distance choices and complex constructions shape persistence, and introducing the \emph{Topological Stability Index (TSI)}, a quantitative measure linking persistence lifetimes to systemic resilience.


\section{Theoretical and Methodological Considerations}
\label{sec:theoretical}

The preceding case studies illustrated how topological structure can reveal hidden organization in business and financial data.  
This section develops the theoretical and methodological foundation behind those observations.  
It focuses on three core layers of the TDA framework: the definition of distance, the construction of simplicial complexes, and the translation of persistence into interpretable indicators.  
While the exposition remains accessible, the goal is to clarify why TDA produces robust and meaningful geometric summaries that complement classical statistical models.

\subsection{Algebraic Formalism of Persistent Homology}\label{homtheory}

Topological Data Analysis translates data into algebraic structure by tracking the evolution of topological features across a sequence of nested spaces. This process is formalized using tools from algebraic topology.

Let $X$ be a finite metric space (e.g., a point cloud or a distance matrix) and $\{\mathcal{K}_\varepsilon\}_{\varepsilon \geq 0}$ a filtration of simplicial complexes constructed from $X$, typically using the Vietoris–Rips complex. For each $\varepsilon$, we compute the $k$-th homology group with coefficients in a field $\mathbb{F}$:
\[
H_k(\mathcal{K}_\varepsilon; \mathbb{F}) = \frac{\ker \partial_k}{\operatorname{im} \partial_{k+1}},
\]
where $\partial_k$ is the boundary operator mapping $k$-simplices to $(k-1)$-simplices. The dimension of $H_k$ is called the $k$-th Betti number $\beta_k$, which counts the number of $k$-dimensional holes (e.g., $\beta_0$ counts components, $\beta_1$ counts loops).

As $\varepsilon$ increases, we obtain a sequence of inclusions:
\[
\mathcal{K}_{\varepsilon_1} \hookrightarrow \mathcal{K}_{\varepsilon_2} \hookrightarrow \cdots,
\]
which induces a sequence of homology group homomorphisms:
\[
H_k(\mathcal{K}_{\varepsilon_1}) \to H_k(\mathcal{K}_{\varepsilon_2}) \to \cdots.
\]
A topological feature (e.g., a loop) is said to be \emph{born} at scale $\varepsilon_b$ if it appears in $H_k(\mathcal{K}_{\varepsilon_b})$ but not earlier, and \emph{dies} at $\varepsilon_d$ if it becomes trivial or merges with an older feature. The collection of such $(\varepsilon_b, \varepsilon_d)$ pairs is the \emph{persistence diagram} $D_k$.

The algebraic stability theorem guarantees that small perturbations in the input data induce small changes in the persistence diagram (with respect to the bottleneck distance), making these summaries robust for noisy empirical data \cite{CohenSteiner2007}.

\subsection{Distances: What Are We Measuring?}

The notion of \emph{distance} is the first, and often most consequential, modeling choice in Topological Data Analysis.  
It determines what kind of similarity the analyst deems meaningful and directly shapes the geometry from which persistence is computed.  
Formally, a distance (or metric) $d(x_i, x_j)$ satisfies non-negativity, identity, symmetry, and the triangle inequality, yet in practice it also serves a semantic role: it encodes the domain’s definition of proximity in behavioral, financial, or operational terms.

\paragraph{Euclidean and Mahalanobis distances.}
For cross-sectional data such as customer profiles, product attributes, or firm-level indicators, the Euclidean metric
\[
d_{ij} = \|x_i - x_j\|_2
\]
remains the most common.  
It measures absolute dissimilarity but assumes that all variables are commensurate.  
When correlations between variables are non-negligible, the Mahalanobis distance introduces covariance normalization, effectively rescaling the space by its variance–covariance structure.  
This adjustment is crucial in finance and econometrics, where multiple factors interact and simple Euclidean proximity may distort relationships.

\paragraph{Correlation and cosine distances.}
For time-dependent or compositional data, distances based on co-movement are more informative.  
The correlation-based distance
\[
d_{ij} = \sqrt{2(1 - \rho_{ij})}
\]
and its directional analogue, the cosine distance, measure similarity in shape rather than magnitude.  
They are invariant to linear rescaling and highlight structural alignment, namely, two assets or behaviors moving in parallel but at different amplitudes remain ``close''.  
This makes them particularly suitable for financial returns, marketing signals or sentiment trajectories.

\paragraph{Dynamic and symbolic distances.}
Sequential data often exhibit time shifts or variable speeds.  
Dynamic Time Warping (DTW) compensates by elastically aligning patterns before computing distances, allowing comparison of behaviors that evolve at different temporal paces.  
In symbolic representations such as SAX or eSAX, distances are computed between symbolic strings using edit or block metrics, capturing coarse-grained similarity while filtering out high-frequency noise.  
Such symbolic metrics emphasize overall behavioral resemblance, though fine local variations may be suppressed.

\paragraph{Interpretation.}
The chosen distance determines the lens through which TDA perceives data structure.  
Selecting it appropriately aligns mathematical abstraction with economic meaning:
\begin{itemize}[leftmargin=1.5em]
  \item In \emph{marketing}, Euclidean or symbolic distances capture similarity in consumer trajectories or purchase patterns.
  \item In \emph{finance}, correlation-based metrics reflect systemic co-movement among assets or indices.
  \item In \emph{macroeconomics}, Wasserstein or earth-mover distances compare distributions of indicators across countries or time.
\end{itemize}

\paragraph{Stability principle.}
A central theoretical result in TDA is the \emph{stability of persistence diagrams} \cite{CohenSteiner2007}.  
Small perturbations in the input distance matrix produce only proportionally small changes, in the bottleneck distance, between the resulting diagrams.  
Consequently, if the chosen distance captures a meaningful notion of similarity, the qualitative features extracted by TDA remain robust to noise, outliers, and sampling uncertainty.  
This property underpins TDA’s reliability and interpretability in empirical analytics, distinguishing it from many model-based techniques whose output can fluctuate under minor data variation.

\subsection{Complexes and Filtrations: Aggregating Local Geometry}

Once a distance measure is fixed, the next step is to translate pairwise proximities into a global geometric object.  
This is achieved through a \emph{simplicial complex}, a collection of vertices, edges, triangles, and higher-dimensional faces that represent how data points cluster and overlap.  
Simplicial complexes transform local similarity information into a combinatorial representation of the data’s shape.

\paragraph{Vietoris-Rips complex.}
The Vietoris-Rips complex $R_\varepsilon$ includes all simplices whose vertices are pairwise within a distance threshold~$\varepsilon$.  
It depends solely on the distance matrix and is therefore applicable even when coordinates are unavailable.  
The Rips complex scales well to high-dimensional or non-Euclidean data, making it the default choice in most empirical TDA applications, including all three case studies presented earlier.

\paragraph{\v{C}ech and Alpha complexes.}
When data points have explicit Euclidean coordinates, more geometrically faithful constructions can be employed.  
The \v{C}ech complex includes a simplex whenever the corresponding $\varepsilon/2$-balls intersect, thus reproducing the exact topology of the underlying union of balls.  
The Alpha complex refines this idea via Delaunay triangulations, producing a smaller but topologically equivalent subcomplex.  
Both constructions offer superior geometric accuracy but are computationally heavier and therefore less common in large-scale analytics.

\paragraph{Witness complexes.}
For very large datasets, the number of simplices in a full Rips complex grows exponentially.  
\emph{Witness complexes} provide an efficient approximation by selecting a representative subset of \emph{landmark} points and using the remaining data as witnesses that determine which simplices to include.  
This strategy preserves essential topological information while drastically reducing computational cost, making it suitable for big-data applications such as consumer-stream analysis or high-frequency trading networks.

\paragraph{Filtrations and multi-scale structure.}
Rather than fixing a single threshold~$\varepsilon$, TDA examines how connectivity evolves as $\varepsilon$ varies.  
This produces a nested sequence of complexes
\[
R_{\varepsilon_1} \subseteq R_{\varepsilon_2} \subseteq \cdots \subseteq R_{\varepsilon_m},
\]
known as a \emph{filtration}.  
Tracking how features appear and disappear along this sequence reveals structure across scales:  
short-lived features correspond to local noise, while long-lived ones represent persistent, global organization.  
Filtrations thus act as a geometric analogue of multi-resolution analysis in signal processing—nonparametric, data-driven and inherently multi-scale.

\subsection{From Persistence to Indicators: Quantifying Structure}

Persistent homology converts the filtration into a set of features with measurable lifetimes.  
For each topological dimension~$k$, the corresponding persistence diagram~$D_k$ contains points $(b_i, d_i)$ representing the birth and death of the $i$-th feature as the scale~$\varepsilon$ increases.  
These diagrams serve as the central output of TDA: they summarize how long each structural pattern (connected component, loop or void) persists across scales.  
Analytically, persistence diagrams can be treated either \emph{qualitatively}, through visual inspection of barcodes or diagrams, or \emph{quantitatively}, through derived numerical summaries.

\paragraph{Total persistence.}
A widely used scalar summary is the \emph{total persistence}
\[
\mathrm{TP}_k = \sum_i (d_i - b_i),
\]
which measures the overall topological activity in dimension~$k$.  
Large total persistence implies rich and long-lived structure, greater heterogeneity, stronger clustering or more pronounced cyclic organization.  
In economic data, a rise in total persistence may reflect regime diversification, market segmentation or systemic instability depending on context.

\paragraph{Lifetime variance and entropy.}
The distribution of lifetimes provides additional information about structural stability.  
The variance or entropy of the lifetimes $\{d_i - b_i\}$ quantifies whether persistence is concentrated in a few dominant features or spread across many transient ones.  
Systems with numerous short-lived loops or components often display erratic, unstable behavior, whereas dominance by a few long-lived features signals consistent structural organization.

\paragraph{Topological similarity.}
To compare persistence diagrams across time, categories, or markets, one can use distances such as the \emph{bottleneck} or \emph{Wasserstein} metrics \cite{Chazal2014}.  
These provide a principled measure of structural proximity between systems, allowing clustering, classification, or trend tracking directly in the topological feature space.  
In this sense, persistence diagrams act as \emph{fingerprints of structure}: objects that can be compared quantitatively while remaining interpretable.

\subsection{The Topological Stability Index (TSI)}

While persistence diagrams encapsulate rich geometric information, practitioners often require concise numerical indicators that can be tracked and compared over time.  
To this end, we introduce the \emph{Topological Stability Index} (TSI), a simple, scale-agnostic measure that quantifies the dispersion of structural lifetimes and thereby reflects the system’s degree of organizational stability.

Let $\mathcal{L}_t$ denote the multiset of feature lifetimes $(d_i - b_i)$ observed in homology dimensions $H_0$ and $H_1$ within a rolling time window~$t$.  
The TSI is defined as
\[
\mathrm{TSI}_t = \operatorname{Var}(\mathcal{L}_t),
\]
that is, the variance of persistence lifetimes within the window.

\paragraph{Interpretation.}
The intuition behind TSI is straightforward:
\begin{itemize}[leftmargin=1.5em]
  \item \textbf{Low TSI} values indicate that all topological features have similar lifetimes, suggesting structural equilibrium, either a stable, well-organized configuration or uniformly random noise.  
  \item \textbf{High TSI} values imply a heterogeneous mix of short- and long-lived features, signaling transitions between order and disorder.  
\end{itemize}
In economic systems, a rising TSI coincides with periods of structural reorganization: increasing systemic stress in financial networks, emerging volatility in consumer segmentation, or shifts in macroeconomic alignments.

\paragraph{Normalization and extensions.}
Because the scale of persistence can vary across datasets, it is useful to normalize TSI by total persistence:
\[
\mathrm{nTSI}_t = \frac{\operatorname{Var}(\mathcal{L}_t)}{\mathrm{TP}_t + \epsilon},
\]
where $\epsilon$ is a small constant preventing division by zero.  
The normalized index~$\mathrm{nTSI}_t$ expresses relative instability—variance of structural lifetimes per unit of total topological activity.  
It can be visualized as a time series, analogous to volatility or entropy measures, thereby linking topology to familiar empirical indicators used in monitoring and risk assessment.

\paragraph{Relation to existing risk measures.}
Conceptually, TSI plays a role analogous to variance in classical statistics, but in the \emph{space of topological lifetimes} rather than numerical returns.  
While volatility measures the dispersion of amplitudes, TSI measures the dispersion of \emph{structural persistence}.  
In portfolio networks, an increase in TSI may reveal diversification breakdowns before they appear in volatility metrics;  
in macroeconomic or trade systems, it can detect early signs of reconfiguration in connectivity patterns or policy alignment.  
TSI thus extends conventional risk analytics from the numerical to the structural domain.

\begin{figure}[H]\centering
\includegraphics[width=0.58\textwidth]{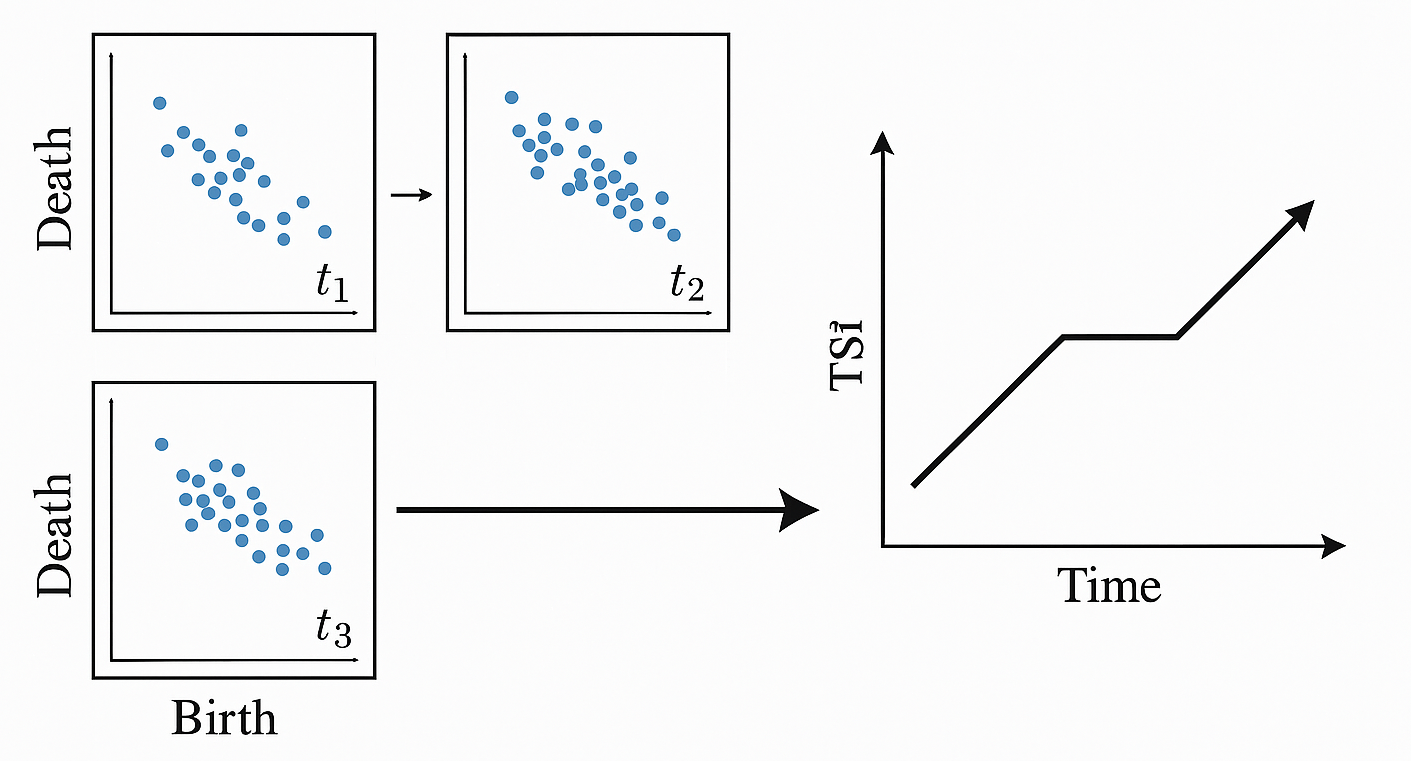}
\caption{Conceptual illustration of the Topological Stability Index (TSI).  
Left: persistence diagrams for three consecutive time windows.  
Right: corresponding TSI values; rising variance of feature lifetimes reflects increasing structural instability and reorganization.}
\label{fig:tsi}
\end{figure}

Figure \ref{fig:tsi} illustrates the conceptual logic behind the Topological Stability Index (TSI).
The left panel shows persistence diagrams for three consecutive time windows $t_1, t_2$ and $t_3$, each representing the system’s topological structure at a given stage.
In these diagrams, each point corresponds to a topological feature (connected component or loop), with its horizontal coordinate marking the \emph{birth} and its vertical coordinate the \emph{death} of the feature.
The vertical distance of a point from the diagonal measures its \emph{lifetime}, i.e., the persistence of that structural feature across scales.

At $t_1$, points are concentrated close to the diagonal, implying that most features appear and disappear rapidly.
This corresponds to a homogeneous or equilibrated structure, where all components behave similarly, resulting in a low TSI.
At $t_2$, a few points begin to separate from the diagonal, representing longer-lived features that coexist with short-lived ones.
The system exhibits partial structural differentiation, and the TSI rises accordingly.
By $t_3$, the dispersion of points is substantial, some features persist for long intervals while others vanish quickly, indicating that the system has become topologically heterogeneous and dynamically unstable.
This stage yields a high TSI, reflecting increased variability in structural lifetimes.

The right panel translates these qualitative changes into a quantitative trajectory:
as the variance of lifetimes across the persistence diagrams grows, the TSI increases over time.
A rising TSI therefore signals the onset of structural reorganization or regime transition.
In financial or economic contexts, such upward trends may precede market instability or systemic stress; in behavioral data, they may correspond to emerging volatility in consumer attention or segmentation dynamics.

A more detailed exploration of the Topological Stability Index, including theoretical properties and empirical calibration, is left for a sequel paper \cite{Diam2026}.

\subsection{Reporting Standards and Reproducibility}

As Topological Data Analysis becomes more prevalent in applied analytics, consistent reporting practices are essential to ensure transparency and reproducibility.
Each empirical study should clearly document the modeling choices that shape its topological results, including: 

\begin{itemize}[leftmargin=1.2em]
\item \textbf{Data and pre-processing:} dataset description, time-windowing scheme, normalization, detrending, and any symbolic transformation applied;
\item \textbf{Distance metric:} definition, rationale for selection, and parameter settings (e.g., correlation, DTW, or symbolic distance);
\item \textbf{Complex construction:} type of complex used (Vietoris-Rips, witness, \v{C}ech, or alpha) and range of filtration thresholds~$\varepsilon$;
\item \textbf{Homology dimensions:} the topological dimensions analyzed ($H_0$, $H_1$, optionally $H_2$) and their domain interpretation;
\item \textbf{Software and computation:} algorithms or libraries employed (e.g., \texttt{Ripser}, \texttt{GUDHI}, \texttt{giotto-tda}) and computational settings;
\item \textbf{Interpretation layer:} visualization or feature extraction method used (barcodes, persistence diagrams, images, or derived indices such as the TSI).
\end{itemize}

A concise summary table may accompany each application, mapping these methodological choices to their analytical consequences.
Such documentation not only facilitates replication and benchmarking but also fosters comparability across domains, allowing financial, behavioral, and macroeconomic studies to speak a common geometric language.
Adopting reproducible standards thus transforms TDA from a mathematical tool into a transparent, evidence-based methodology for modern data analytics.

\subsection{Conceptual Summary}

Topology offers a language for describing structure without assuming a specific model.
The stability of persistence guarantees robustness against noise; the filtration framework captures how organization evolves across scales; and derived indices such as the Topological Stability Index (TSI) translate geometric variability into interpretable diagnostics.
Together, these principles explain why TDA generalizes naturally across consumer, market and macro-financial domains.
They also reconcile mathematical abstraction with empirical insight, positioning topology as both a theoretical foundation and an operational framework for analyzing complexity in economic and behavioral systems.

\section{Practical Guidelines and Reporting Standards}

\subsection{Bridging Mathematics and Practice}

While the conceptual foundations of TDA are mathematically rigorous, its impact in applied analytics depends on transparent, interpretable and reproducible implementation.
This section provides concrete guidelines for analysts and researchers applying TDA in business, financial or behavioral settings.
The focus is on clarity and methodological consistency rather than algorithmic novelty, aiming to bridge the gap between formal theory and operational practice.

\subsection{A Minimal Reporting Checklist}

Table~\ref{tab:checklist} summarizes essential information that should accompany every empirical TDA study.  
Such transparency enables replication, facilitates comparison across domains, and ensures that structural interpretations are grounded in coherent methodological choices.

\begin{table}[H]
\centering
\caption{Minimal reporting checklist for empirical TDA studies.}
\label{tab:checklist}
\begin{tabular}{@{}p{0.25\linewidth}p{0.65\linewidth}@{}}
\toprule
\textbf{Component} & \textbf{Information to Report and Justify} \\
\midrule
\textbf{Data Source and Scope} & Dataset origin, frequency, time span, variables, pre-processing (normalization, detrending, missing-data handling). \\
\textbf{Distance Metric} & Definition and rationale (Euclidean, correlation, cosine, DTW, symbolic, etc.). Clarify why it matches domain semantics. \\
\textbf{Complex Type} & Vietoris–Rips, \v{C}ech, Alpha or Witness complex; filtration range. \\
\textbf{Persistence Dimensions} & Which homology dimensions are analyzed ($H_0$, $H_1$, possibly $H_2$) and why these are meaningful. \\
\textbf{Software and Parameters} & Library used (e.g., \texttt{GUDHI}, \texttt{Ripser}, \texttt{giotto-tda}), numerical precision and computational settings. \\
\textbf{Summary Statistics} & Total persistence, lifetime variance, Topological Stability Index (TSI), or other derived indicators. \\
\textbf{Interpretation Framework} & Economic or managerial meaning of persistent features (clusters, cycles, stability). \\
\textbf{Validation} & Sensitivity analysis across metrics, window sizes and filtration parameters. \\
\bottomrule
\end{tabular}
\end{table}

\subsection{Good Practices for Applied Analysts}

The following recommendations enhance the robustness, interpretability, and credibility of applied TDA studies:

\begin{itemize}[leftmargin=1.2em]
  \item \textbf{Normalize before geometry.} Ensure that the distance matrix reflects genuine structure, not scale differences or non-stationarity.
  \item \textbf{Prefer simple complexes.} Vietoris–Rips complexes are sufficient in most business and financial settings; only move to Alpha or Witness when computationally necessary.
  \item \textbf{Track persistence over time.} Rolling-window analyses reveal when topological features appear or vanish, linking them to economic or behavioral events.
  \item \textbf{Visualize with clarity.} Use persistence diagrams and barcodes alongside standard plots (correlation matrices, PCA) to highlight complementary insights.
  \item \textbf{Quantify uncertainty.} Report sensitivity of persistence features to metric choice and noise; bootstrap diagrams if feasible.
  \item \textbf{Integrate with domain metrics.} Integrate TDA outputs with volatility indices, sentiment measures, or KPIs to strengthen managerial relevance.
\end{itemize}

\subsection{Reproducibility and Open Data}

To encourage cumulative progress, reproducible scripts and open datasets should accompany all TDA studies whenever licensing allows.  
Providing persistence diagrams, distance matrices, and parameter files facilitates comparison and benchmarking.  
Repositories such as GitHub or institutional archives enable transparent dissemination and foster interdisciplinary reuse, bridging mathematics, data science and applied economics.

\subsection{From Workflow to Standard Practice}

Over time, TDA can become part of the standard analytics toolbox, comparable to correlation or PCA today.  
For this to occur, two practices are crucial: (i) explicit documentation of methodological choices, and (ii) consistent reporting of interpretive conventions (e.g., what constitutes a ``long-lived'' feature).  
Adherence to such conventions will allow future meta-analyses, enabling systematic assessment of topological indicators across markets and behavioral datasets.

\section{Conclusion and Outlook}

This work has shown how Topological Data Analysis can function as a unifying geometric framework for business, financial and behavioral analytics.
By focusing on the \emph{shape} of data rather than its point-wise fluctuations, TDA reveals forms of structural organization that complement, and often extend, traditional statistical tools.

\smallbreak 

Across the three comparative studies of \S~\ref{sec:case}, persistent homology captured:
\begin{itemize}[leftmargin=1.2em]
  \item sectoral cohesion and cyclical co-movements in equity markets;
  \item recurring waves of collective attention in consumer behavior;
  \item regional and policy-aligned currency blocs in the foreign-exchange network.
\end{itemize}
Despite the diversity of domains, these applications share a common geometric logic: stable topological features correspond to persistent economic organization, while rapid changes in persistence signal transitions, volatility or structural reconfiguration.

The theoretical analysis clarified why this framework is both robust and interpretable.
Persistence diagrams are mathematically stable under noise, scale naturally across thresholds, and summarize connectivity and cycles in ways that remain meaningful across contexts.
The proposed \emph{Topological Stability Index} (TSI) translates these abstract notions into quantitative diagnostics for systemic risk, market dynamics and consumer behavior, bridging topological insight with managerial and policy relevance.

\paragraph{Future Directions.}
The next frontier for applied TDA lies in integration across analytical paradigms and domains:
\begin{enumerate}[leftmargin=2em]
\item \textbf{Hybrid modeling:} combining symbolic, statistical and topological features within machine-learning pipelines for forecasting, segmentation and anomaly detection;
\item \textbf{Dynamic systems:} extending persistence analysis to evolving networks and non-stationary manifolds, enabling real-time monitoring of market or behavioral transitions;
\item \textbf{Cross-domain applications:} applying topology to sustainability analytics, consumer sentiment networks, innovation diffusion and policy-coordination studies.
\end{enumerate}

\paragraph{Closing reflection.}
Modern data environments are increasingly relational, nonlinear and dynamic.
In such settings, topology provides a coherent language for describing stability, transition and emergence.
By linking mathematical structure to empirical meaning, TDA bridges theory and application, offering a scalable, interpretable paradigm for understanding the hidden geometry of data.

\medskip
\noindent\textbf{Acknowledgements.}  
The author gratefully acknowledges the contributions of his students Fredrik Hobbelhagen, Pola Bereta and Pattravadee (Marie) de Favereau de Jeneret, whose research inspired and informed the comparative studies presented in this survey.

\end{document}